\documentclass[journal]{IEEEtran}
\usepackage{amsmath,amsfonts}
\usepackage[noend,ruled,linesnumbered]{algorithm2e}
\usepackage{booktabs}
\usepackage{multirow}
\usepackage{tabularx}
\usepackage{array}
\usepackage[caption=false,font=normalsize,labelfont=sf,textfont=sf]{subfig}
\usepackage{textcomp}
\usepackage{url}
\usepackage{verbatim}
\usepackage{graphicx}
\usepackage{cite}
\usepackage{ragged2e}
\usepackage{hyperref}
\usepackage{subfloat}
\usepackage{color}
\newtheorem{definition}{Definition}
\usepackage{tikz,xcolor,hyperref}
\definecolor{lime}{HTML}{A6CE39}
\DeclareRobustCommand{\orcidicon}{%
    \begin{tikzpicture}
    \draw[lime, fill=lime] (0,0) 
    circle [radius=0.16] 
    node[white] {{\fontfamily{qag}\selectfont \tiny ID}};    \draw[white, fill=white] (-0.0625,0.095) 
    circle [radius=0.007];    \end{tikzpicture}
    \hspace{-2mm}}
\foreach \x in {A, ..., Z}{%
    \expandafter\xdef\csname orcid\x\endcsname{\noexpand\href{https://orcid.org/\csname orcidauthor\x\endcsname}{\noexpand\orcidicon}}
    }

\begin{document}

\title{UAV-assisted Joint Mobile Edge
Computing and Data Collection via Matching-enabled Deep Reinforcement Learning}

\author{Boxiong Wang\orcidA{}, 
Hui Kang\orcidB{}, 
Jiahui Li\orcidC{}, 
Geng Sun\orcidD{}, \IEEEmembership{Senior Member, IEEE,}
Zemin Sun\orcidE{}, 
Jiacheng Wang\orcidF{},\\
and Dusit Niyato\orcidG{},~\IEEEmembership{Fellow, IEEE}
\thanks{This work is supported in part by the National Natural Science Foundation of China (62172186, 62272194, 62471200), in part by the Science and Technology Development Plan Project of Jilin Province (20240302079GX), in part by the National Research Foundation, Singapore, in part by Infocomm Media Development Authority under its Future Communications Research \& Development Programme (FCP-NTU-RG-2022-010 and FCP-ASTAR-TG-2022-003), in part by Singapore Ministry of Education (MOE) Tier 1 (RG87/22 and RG24/24), in part by the NTU Centre for Computational Technologies in Finance (NTU-CCTF), in part by the RIE2025 Industry Alignment Fund - Industry Collaboration Projects (IAF-ICP) (Award I2301E0026), administered by A*STAR, in part by Alibaba Group and NTU Singapore through Alibaba-NTU Global e-Sustainability CorpLab (ANGEL), in part by the Postdoctoral Fellowship Program of China Postdoctoral Science Foundation (GZC20240592), in part by the China Postdoctoral Science Foundation General Fund (2024M761123), and in part by the Scientific Research Project of Jilin Provincial Department of Education (JJKH20250117KJ). \textit{(Corresponding authors: Jiahui Li and Geng Sun.)}
\par Boxiong Wang, Hui Kang, Jiahui Li, and Zemin Sun are with the College of Computer Science and Technology, Jilin University, Changchun 130012, China. Hui Kang is also with the Key Laboratory of Symbolic Computation and Knowledge Engineering of Ministry of Education, Jilin University, Changchun 130012, China. (E-mails: wangbx0320@163.com; lijiahui@jlu.edu.cn; sunzemin@jlu.edu.cn; kanghui@jlu.edu.cn).
\par Geng Sun is with the College of Computer Science and Technology, Jilin University, Changchun 130012, China, and also with the Key Laboratory of Symbolic Computation and Knowledge Engineering of Ministry of Education, Jilin University, Changchun 130012, China. He is also with the College of Computing and Data Science, Nanyang Technological University, Singapore 639798 (E-mail: sungeng@jlu.edu.cn).
\par Jiacheng Wang and Dusit Niyato are with the College of Computing and Data Science, Nanyang Technological University, Singapore (E-mails: jiacheng.wang@ntu.edu.sg; dniyato@ntu.edu.sg).
}
}



\maketitle

\begin{abstract}
\par Unmanned aerial vehicle (UAV)-assisted mobile edge computing (MEC) and data collection (DC) have been popular research issues. Different from existing works that consider MEC and DC scenarios separately, this paper investigates a multi-UAV-assisted joint MEC-DC system. Specifically, we formulate a joint optimization problem to minimize the MEC latency and maximize the collected data volume. This problem can be classified as a non-convex mixed integer programming problem that exhibits long-term optimization and dynamics. Thus, we propose a deep reinforcement learning-based approach that jointly optimizes the UAV movement, user transmit power, and user association in real time to solve the problem efficiently. Specifically, we reformulate the optimization problem into an action space-reduced Markov decision process (MDP) and optimize the user association by using a two-phase matching-based association (TMA) strategy. Subsequently, we propose a soft actor-critic (SAC)-based approach that integrates the proposed TMA strategy (SAC-TMA) to solve the formulated joint optimization problem collaboratively. Simulation results demonstrate that the proposed SAC-TMA is able to coordinate the two subsystems and can effectively reduce the system latency and improve the data collection volume compared with other benchmark algorithms.
\end{abstract}

\begin{IEEEkeywords}
UAV communications, MEC, data collection, DRL, resource allocation.
\end{IEEEkeywords}


\section{Introduction}

\par The advancement of wireless networks and the development of Internet of Things (IoT) manufacturing technology have facilitated the exponential growth of IoT applications, which are now ubiquitous in a multitude of domains, including industry, transportation, environmental monitoring, modern smart city applications, and agriculture \cite{Adnan2024, ZheSurvey}. For example, IoT devices such as wireless sensors, are often deployed on the ground and are able to detect and sense the environment to assist in tasks such as factory inspection, traffic management, and environmental monitoring. Nevertheless, IoT devices typically have limited computing capability as well as storage resources, hence they need to transmit data through wireless networks to nearby base stations or data centers for processing. In traditional wireless networks, IoT devices rely on static nodes to provide services and usually transmit sensed data through multi-hop relays~\cite{Zhu2024}. However, in some remote scenarios such as factory safety inspections and livestock monitoring in farmlands, IoT devices are usually deployed in hard-to-reach places. Consequently, these IoT devices encounter significant challenges in uploading their data to remote access points or data centers \cite{Du2024}. 

\par In this case, unmanned aerial vehicles (UAVs) can play a crucial role in IoT due to their mobility, low cost, and rapid deployment. Specifically, UAVs can assist in wireless networks, which means that they can fly to the hard-to-reach places to provide services as the aerial base station~\cite{huang2024a,zhang2024,zhang2024a}. Furthermore, UAVs can hover over IoT devices to establish line-of-sight (LoS) links, thereby significantly improving the quality of communication for IoT devices \cite{Wu2021,Jiang2023}. Therefore, UAVs can assist wireless networks and are a promising technology in IoT. Generally, UAV technology is widely applied in two scenarios in IoT, which are UAV-assisted mobile edge computing (MEC) and UAV-assisted data collection (DC), respectively. In these two scenarios, UAVs can provide edge computing services with high LoS probability that improve coverage and quality of service (QoS) of wireless networks and can act as mobile data sink nodes to collect data directly from IoT devices, thereby reducing transmit power requirements of IoT devices and preventing data overflow. Moreover, in some scenarios such as agricultural management \cite{pei2022,Zhou2022a}, traffic management \cite{Masuduzzaman2022,Elloumi2018}, and post-disaster relief \cite{sun2024,dong2021}, there are both computation-intensive tasks and DC requirements, requiring multiple UAVs to perform MEC and DC simultaneously. For example, UAVs collect data and then perform edge computing or offload the data to a nearby data center.

\par However, due to the energy constraints of UAVs and priority differences of tasks, the UAV-assisted MEC and UAV-assisted DC need to be performed in different UAVs, and they are usually studied separately (e.g., \cite{zheng2024,sun2024a,sun2024b,Li2024a,liu2024b,Sun2023}). Specifically, some MEC tasks with hard deadlines usually require real-time and continuous computing support. If an MEC-UAV performs DC simultaneously, it may lead to a decline in the real-time performance of the MEC tasks and higher energy consumption. Moreover, UAVs from different operators might carry out different tasks in the same area, such as edge computing within a factory and environmental monitoring around the factory, or post-disaster rescue and post-disaster DC for evaluation. Furthermore, although there has been some research on data sharing security and privacy protection in edge computing systems, such as the privacy-aware and secure matching encryption (PS-ME) method and the privacy-preserving fine-grained data sharing scheme with dynamic service (PF2DS) proposed by Sun et al \cite{sun2023a,sun2024privacy}, isolating MEC and DC data onto different UAVs remains one of the effective means to protect data privacy and security.

\par Constructing such an abovementioned joint MEC-DC system has several major challenges. First, the shortage of spectrum resources is a growing problem in current and future networks, which means that the spectrum-sharing techniques become particularly important. In this context, the issue of co-channel interference needs to be addressed, while the existing UAV-assisted communication works usually assume that the interference is ignored (e.g.,~\cite{Li2024,Chen2023,Dandapat2024,Zhao2024,Yang2020}). Conversely, when considering interference, the uplink communication of UAV-assisted MEC and UAV-assisted DC will be simultaneously affected. Second, due to the limited energy constraints, UAVs require proper trajectory optimization to ensure the continuous operation of the network. Moreover, given the limited transmit power of IoT devices, i.e., ground users (GUs) and the limited computing and storage resources carried by UAVs, it is necessary to prioritize the most suitable users for service to enhance the efficiency of the system. Therefore, careful consideration of the abovementioned issues is required to improve the quality of both MEC and DC.

\par However, finding an appropriate solution during MEC and DC while jointly considering the interference, energy constraints, user association, UAV trajectories, and user transmit power is a challenging task. For example, it is possible to maximize the performance of the MEC subsystem by covering as many GUs as possible or adjusting the user transmit power of MEC, while this can cause significant interference not only among the MEC-UAVs, but also with the DC-UAVs. Conversely, when DC-users adjust the association and user transmit power, they may cause serious interference to MEC-UAVs. To mitigate this issue, UAVs can fly to better positions to reduce the impact of interference. However, this could lead to excessive energy consumption, potentially exceeding energy constraints. Therefore, the goals of MEC and DC are conflicting and interdependent within the same scenario and are difficult to balance. Different from previous works that only considering the optimization of separate MEC or DC systems, this paper proposes a joint optimization approach to improve the performance of MEC and DC systems simultaneously. The main contributions of this paper are summarized as follows.

\begin{itemize}
    \item \textit{Joint MEC and DC System:} We consider a multi-UAV-assisted joint MEC-DC system to coordinate the UAVs to perform MEC and DC simultaneously. Specifically, this system combining multiple UAVs for processing computation-intensive MEC tasks and a single UAV for freshness-insensitive DC, which considers the co-channel interference among UAVs. To the best of our knowledge, such a joint MEC-DC system with mutual effect has not yet been investigated in the literature.
    
    \item \textit{Joint Optimization Problem Formulation:} Regarding the two objectives of the total system latency of MEC and the data volume of DC, we find they are conflicting with each other. Accordingly, we formulate a joint optimization problem that aims to minimize the total system latency of MEC while maximizing the total volume of collected data simultaneously by adjusting the movement of UAVs, user association, and transmit power of the users. Moreover, this optimization appears to be a mixed-integer non-convex problem with dynamic and long-term optimization properties.
    
    \item \textit{Deep Reinforcement Learning (DRL)-based Approach Design:} We propose a DRL-based approach to solve the optimization problem effectively. Specifically, we reformulate the problem into an action space-reduced Markov decision process (MDP) by modeling the user association as a one-to-many matching game with externalities. Based on this, we propose an SAC-based algorithm integrated with a two-phase matching-based association (TMA) strategy to optimize the UAV movement, user association, and user transmit power in real time to jointly enhance the system performance of both MEC and DC.
    
    \item \textit{Performance Evaluation and  Analysis:} Simulation results demonstrate that the proposed SAC-TMA algorithm outperforms the other four benchmark algorithms under different numbers of MEC-users, which can learn effective policies to jointly enhance the latency, collected data volume, and coverage performance of MEC and DC. Besides, our proposed TMA strategy outperforms traditional matching-based algorithms and random strategies, and it is feasible in terms of algorithm running time.
\end{itemize}

\par The rest of this paper is arranged as follows. Section~\ref{sec:related works} reviews the related work. In Section \ref{sec:system_model}, the system models are presented. The joint optimization problem and formulation are presented in Section \ref{sec:problem formulation}. Section \ref{sec:proposed algorithm} proposes the SAC-TMA and Section \ref{sec:experiments_and_analysis} provides the simulation results and analysis. Finally, the overall paper is concluded in Section~\ref{sec:conclusion}.

\begin{table*}[htb]
    \scriptsize
    \centering
    \caption{Differences between This Work and Existing Works}
    \label{tab:works comparison}
    \begin{tabular*}{\textwidth}{@{}@{\extracolsep{\fill}}cccccccc@{}}
\toprule
& \textbf{Reference} & \begin{tabular}[c]{@{}c@{}}\textbf{Multi-UAV} \\ \textbf{system} \end{tabular} & \begin{tabular}[c]{@{}c@{}}\textbf{Single-UAV} \\ \textbf{system} \end{tabular} & \begin{tabular}[c]{@{}c@{}}\textbf{Real-time} \\ \textbf{tasks}\end{tabular} & \begin{tabular}[c]{@{}c@{}}\textbf{Consider}\\ \textbf{Interference}\end{tabular} & \begin{tabular}[c]{@{}c@{}}\textbf{Long-term} \\ \textbf{optimization}\end{tabular} & \begin{tabular}[c]{@{}c@{}}\textbf{Online} \\ \textbf{algorithm}\end{tabular} \\ \hline \hline
\multicolumn{1}{c}{\multirow{9}{*}{\begin{tabular}[c]{@{}c@{}}UAV-Assisted\\ MEC Systems\end{tabular}}}    & \cite{Yu2020}  & $\times$            & $\checkmark$      & $\times$    & $\times$      & $\times$               & $\times$      \\
\multicolumn{1}{c}{}          & \cite{Zhan2021}  & $\checkmark$                & $\times$  & $\times$    & $\checkmark$      & $\times$               & $\times$      \\
\multicolumn{1}{c}{}          & \cite{Zhou2022}  & $\times$                & $\checkmark$      & $\times$    & $\times$      & $\times$               & $\times$      \\
\multicolumn{1}{c}{}          & \cite{Chen2023}  & $\checkmark$                & $\times$  & $\checkmark$    & $\times$      & $\checkmark$               & $\checkmark$      \\
\multicolumn{1}{c}{}          & \cite{Li2024}  & $\checkmark$            & $\times$      & $\checkmark$    & $\times$      & $\checkmark$               & $\checkmark$          \\
\multicolumn{1}{c}{}          & \cite{Lee2024}  & $\checkmark$            & $\times$      & $\checkmark$    & $\times$      & $\checkmark$           & $\checkmark$     \\ 
\multicolumn{1}{c}{}          & \cite{Pervez2024}  & $\checkmark$            & $\times$      & $\times$    & $\times$      & $\times$           & $\times$     \\ 
\multicolumn{1}{c}{}          & \cite{wang2023a}  & $\times$            & $\checkmark$      & $\checkmark$    & $\times$      & $\checkmark$           & $\checkmark$     \\  
\multicolumn{1}{c}{}          & \cite{Liu2022}  & $\checkmark$            & $\times$      & $\checkmark$    & $\times$      & $\checkmark$           & $\checkmark$     \\ \hline\hline
\multicolumn{1}{c}{\multirow{5}{*}{\begin{tabular}[c]{@{}c@{}}UAV-Assisted \\ DC Systems\end{tabular}}}     & \cite{Yu2021}  & $\times$            & $\checkmark$      & $\checkmark$    & $\times$  & $\checkmark$           & $\checkmark$     \\
\multicolumn{1}{c}{}              & \cite{Dandapat2024}  & $\checkmark$                & $\times$      & $\times$    & $\times$  & $\times$           & $\times$     \\
\multicolumn{1}{c}{}              & \cite{Zhu2024}  & $\times$            & $\checkmark$  & $\checkmark$    & $\times$  & $\checkmark$       & $\checkmark$   \\
\multicolumn{1}{c}{}              & \cite{Du2024}  & $\checkmark$            & $\times$      & $\times$        & $\checkmark$      & $\times$           & $\times$     \\
\multicolumn{1}{c}{}             & \cite{Liu2024}  & $\times$        & $\checkmark$  & $\times$    & $\checkmark$      & $\times$           & $\times$         \\  \hline\hline
\multicolumn{1}{c}{\multirow{3}{*}{\begin{tabular}[c]{@{}c@{}}UAV-Assisted\\ Hybrid Systems\end{tabular}}} & \cite{Zeng2023}  & $\times$         & $\checkmark$      & $\checkmark$        & $\times$  & $\times$      & $\times$          \\
\multicolumn{1}{c}{}             & \cite{Liu2024a}  & $\checkmark$         & $\times$  & $\checkmark$    & $\times$      & $\checkmark$       & $\checkmark$               \\
\multicolumn{1}{c}{}             & \textbf{This work} & $\checkmark$  & $\checkmark$    & $\checkmark$  & $\checkmark$       & $\checkmark$  & $\checkmark$ \\ \bottomrule
        \end{tabular*}
\end{table*}

\section{Related Works}\label{sec:related works}

\par In this section, we review research on UAV-assisted hybrid MEC-DC architecture, joint optimization for MEC and DC, and optimization approaches. Moreover, Table~\ref{tab:works comparison} illustrates the differences between the state-of-the-art works and this work.

\subsection{UAV-Assisted Hybrid MEC-DC Architecture}

\par UAVs have been widely applied to assist MEC services in some scenarios. For example, Yu \textit{et al.} \cite{Yu2020} studied the optimization problem of collaborative services on UAV and edge clouds, and they proposed a system to control a UAV to provide MEC service in areas where existing edge clouds are inaccessible to IoT devices. Moreover, Zhan \textit{et al.} \cite{Zhan2021} developed a framework for a multi-UAV-assisted MEC system, where multiple UAVs with edge servers offer flexible computing support to IoT devices with time-sensitive requirements. Due to their flexibility and mobility, UAVs can directly collect data from GUs by flying close to them and have drawn significant attention from researchers. For instance, Dandapat \textit{et al.} \cite{Dandapat2024} studied a multi-UAV-assisted DC network, optimizing the three-dimensional (3D) trajectory of the UAVs as well as resource allocation for DC from mobile nodes. Moreover, Liu \textit{et al.} \cite{Liu2024} investigated the trajectory optimization problem of a UAV performing DC tasks in an area containing multiple monitoring regions and multiple base stations in a UAV-assisted DC system. 

\par Previous works on UAV-assisted MEC or UAV-assisted DC primarily focused on separate studies while ignoring the requirements for the scenarios containing both MEC and DC users. Specifically, in real-world scenarios, there may not only compute-intensive tasks to perform, but also large amounts of stored data that require UAV to perform additional data collection tasks. For example, in a smart city environment, UAVs may be required to assist with real-time video analysis for traffic management while simultaneously gathering sensor data from distributed IoT devices, such as air quality sensors, temperature monitors, or noise detectors. Therefore, the UAV assisted hybrid MEC-DC scenarios need further exploration.

\par There have been some studies involving both MEC and DC. For example, Zeng \textit{et al.} \cite{Zeng2023} investigated the UAV-assisted DC and MEC scenario, constructed a new theoretical model for DC rate, and defined the quality of requirement (QoR) for real-time processing. By optimizing the UAV trajectory, resource allocation and task duration, while meeting quality of service and UAV mobility constraints, they were able to reduce the energy consumption of the UAV and task completion time. However, in this work, the UAV only supports data acquisition and relies on nearby MEC servers to fulfill computational requirements. Liu \textit{et al.}~\cite{Liu2024a} investigated a space-air-ground power IoT system and proposed a UAV-enabled wireless power transfer (WPT) framework, where UAVs transfer energy to devices for DC via WPT, utilize MEC for data processing, and eventually forward the data to low earth orbit satellites. Subsequently, they aimed to minimize the average age of information (AoI) of devices by optimizing the number of UAV hovering positions, hovering locations, UAV-device connections, energy transmission, DC time, UAV computational resources, flight speed, and trajectory. However, in~\cite{Liu2024a}, both DC and MEC are completed on the same UAV, and the collected data is actually the data needed for task offloading, which did not consider the conflict between separate MEC and DC systems. Moreover, neither \cite{Zeng2023} nor \cite{Liu2024a} considered the co-channel interference among UAVs.

\par In summary, existing works mainly focused on separate MEC or DC, or collecting data from the same device for MEC, few studies investigated the case where different GUs need to perform MEC and DC separately in the same scenario, which can coordinate the resource allocation of different vendors and achieve privacy protection of data. In this case, the main challenges are the interaction of the MEC and DC subsystems in hybrid scenarios and the UAV trajectory control in the presence of interference among multiple subsystems. This motivates us to investigate these effects and propose an effective approach.

\subsection{Joint Optimization for MEC and DC}

\par Due to the limited computing and storage resources carried by UAVs, resource allocation has been extensively studied to improve the system performance of wireless networks. For example, Wang \textit{et al.}~\cite{wang2023a} investigated the most efficient placement of UAV, resource allocation, and computation offloading to minimize the total delay. Du \textit{et al.} \cite{Du2024} studied a UAV-assisted WPT and DC network and optimized the trajectories of two UAVs, the flight speed, the safe distance of the UAVs, and the energy constraints of each IoT device to increase the minimum DC throughput of the IoT devices. Moreover, Liu \textit{et al.}~\cite{Liu2022} investigated a two-layer UAV-assisted maritime communication network. They proposed a DRL-based approach to reduce the communication and computation latency. However, these studies only considered separate optimization objectives such as MEC latency or DC throughput. 

\par There are other works that studied joint optimization of multiple objectives. For instance, Yu \textit{et al.} \cite{Yu2021} investigated the potential of UAV-assisted wireless-powered IoT network and proposed an extended deep deterministic policy gradient (DDPG) algorithm. Their aim was to achieve a joint optimization objective that maximizes the total energy and data transmission rate while reducing the energy consumption of the UAV. Chen \textit{et al.}~\cite{Chen2023} studied a multi-UAV-aided MEC network and optimized the UAV movement, user association, and user transmit power to jointly minimize the energy consumption and system latency.

\par However, the optimization goals in these works are not suitable for the considered scenario since they did not study the joint optimization objectives of MEC and DC. Thus, they are not capable of solving the challenge of the trade-off between MEC and DC caused by the interference. This motivates us to jointly minimize the MEC latency and maximize the data volume of DC.

\subsection{Optimization Approaches}

\par To solve complex optimization problems, researchers have been dedicated to the design of effective algorithms. For instance, Pervez \textit{et al.}~\cite{Pervez2024} investigated the joint optimization of energy and delay in a UAV-assisted MEC network and proposed a novel three-tier segment-by-segment optimization scheme based on block descent method, simplistic geometric waterfilling, and gradient descent method to solve the problem. However, traditional optimization and heuristic approaches usually require high computational complexity and are difficult to adapt to large-scale and real-time application scenarios, especially when the communication environment changes significantly with the environmental dynamics.

\par To mitigate this issue, DRL-based methods have been extensively studied as an alternative. For instance, Lee \textit{et al.}~\cite{Lee2024} considered a multi-UAV-MEC network and proposed an independent proximal policy optimization (IPPO) model for learning task offloading and trajectory control of UAVs. In~\cite{Zhou2022}, the authors proposed a SAC-based algorithm that maximizes the computation amount and fairness of terminals in a UAV-assisted WPT and MEC system. Li \textit{et al.}~\cite{Li2024} investigated a three-tier multi-UAV-assisted MEC system with random task arrivals, and they proposed a new heterogeneous federated multi-agent reinforcement learning framework, which jointly optimizes task offloading, UAV trajectories, and resource allocation to minimize the AoI. However, the existing DRL-based approaches above did not consider interference among UAVs, which can lead to significant differences in the channel environment. To investigate this effect, Seid \textit{et al.}~\cite{seid2021} considered the inter-cell interference among UAVs and proposed an approach based on multi-agent deep reinforcement learning (MADRL) to ensure the QoS requirements of IoT devices or users while reducing the total computing cost of their considered network. However, this work assumes that UAV clusters provide MEC services to users at fixed positions and does not optimize UAV trajectories. 

\par To sum up, while these works can deal with resource allocation and UAV trajectory optimization in MEC or DC systems, they are not suitable for the considered scenario. The main challenges is the hybrid solution space introduced by the discrete multi-subsystem user association variables and continuous variables and the coupling of decision variables caused by co-channel interference, making it difficult to jointly optimize these variables, especially when optimizing the UAV trajectory. Therefore, this prompts us to propose an effective online optimization approach with low computational complexity to address the considered joint optimization problem.

\begin{table*}[]
\caption{List of main notations}
\label{tab:notation}
\begin{tabularx}{\textwidth}{ll}
\toprule
\textbf{Notation}                      & \textbf{Description}                                                                                                                                               \\ \hline
\multicolumn{2}{c}{System model}                                                                                                                                                                          \\ \hline
$g, N_g, \mathcal{G}$                      & The index, number, and set of GUs                                                                                                                                  \\
$m, M, \mathcal{G}^{MEC}$                        & The index, number, and set of MEC-GUs                                                                                                                              \\
$n, N, \mathcal{G}^{DC}$                       & The index, number, and set of DC-GUs                                                                                                                              \\
$f_m(t), f_g(t), F, \mathcal{F}$                        & The task generated by MEC-GU $m$, GU $g$, the number, and set of tasks                                                                                                                                \\
$u, N_U+1, \mathcal{U}$                      & The index, number, and set of UAVs                                                                                                                                 \\
$i, N_U, \mathcal{U}^{MEC}$                      & The index, number, and set of MEC-UAVs                                                                                                                             \\
$u_{dc}$                                        & The index of the DC-UAV                                                                                                                                                         \\
$b_{m,f}, l_{m,f}, t_{m,f}^{max}$                 & The task completion status, number of data bits, and maximum tolerance time limit of the MEC task $f_m(t)$ \\
$D_{m,f}$                                  & The length of deadline for task $f_m(t)$  \\
$\tau, t, T, \mathcal{T}$                  & The length, index, number, and set of time step                                                                                                                     \\
$m(t), \alpha(t)$                          & The distance of movement and the angle of deviation                                                                                                                \\
$V_u(t), V_g(t)$                           & The coordinates of the UAV $u$ and the GU $g$                                                                                                                          \\
$P_{u,g}^{LoS}(t), P_{u,g}^{NLoS}(t)$ & The connection probability of LoS and NLoS  \\
$h_{u,g}(t)$                               & The channel gain between the UAV $u$ and the GU $g$ at time step $t$                                                                                                     \\
$p_g(t), p_u^c, p_g^{max}$                 & The transmit power of the GU $g$ at time step $t$, the computation power of UAV $u$, the maximum power of GU $g$                                                           \\
$W$                                        & Total bandwidth of each UAV                                                                                                                                        \\
$C_i, \omega_i, \kappa_i$                            & The computation intensity, the CPU operating frequency, 
               and the effective switching capacitance on MEC-UAV $i$           \\
$\delta_g$                                 & Task density coefficient       \\
$X_{u,g}(t)$                      & The association indicator variable to represent whether GU $g$ is associated with UAV $u$ at time step $t$                                                        \\
$R_{u,g}(t) $                              & The data transmission rate of GU $g$ associated with UAV $u$                                                                                                           \\
$T_{i,m}^f(t), T_{i,m,f}^c(t), T_i(t)$     & The transmission and computation delay of the task $f_m(t)$ of MEC-user $m$ with MEC-UAV $i$, the total delay of\\
& MEC-UAV $i$ at time step $t$                                                   \\ 
$\mathcal{C}^{MEC}, \mathcal{C}^{DC}$      & The MEC task completion rate and the DC rate\\\hline
\multicolumn{2}{c}{Problem formulation, algorithm, and simulation}                                                                                                                                         \\  \hline
$N_u^{max}$                                & The maximum number of GUs that UAV $u$ can serve                                                                                                                     \\
$D(t), D_{min}, D_{m,f}(t), L_n^{max}$                 & The amount of collected data at time step $t$, minimum amount of data to start collecting, remaining processing time of\\
& the earliest unfinished task $f_m(t)$ of MEC-GU $m$ at time step $t$, data storage limit of the GU $n$                                            \\
$R_{M_{th}}, R_{D_{th}}$                   & Threshold rates for MEC and DC                                                                                                                                     \\
$\sigma, \rho, \delta_p$  & The coefficient in the DC reward function, and the penalty reward for UAVs out of bounds and exceeding \\
                                           & power consumption limits \\
$\varrho, \varsigma$               & The penalty reward variable of collision and the soft update constant                                               \\
$r_l(t), r_d, r_p$                         & The latency reward, DC reward, and penalty reward                                                                                                                  \\
$N_i^f, N_m^f$                             & The number of total tasks completed by the MEC-UAV $i$, the number of total tasks generated by the MEC-user $m$                                                            \\
$L_n(t), l_n(t)$                           & The amount of data storage for the DC-user $n$, the data volume generated by the DC-user $n$ at time step $t$                                                                               \\
\bottomrule
\end{tabularx}%
\end{table*}

\section{System Model}
\label{sec:system_model}

\par In this section, we present the adopted models in the considered UAV-assisted joint MEC-DC network, and the main notations are listed in Table \ref{tab:notation}. 

\subsection{Network Model}
\begin{figure}[tbp]
\centering
\includegraphics[width=0.47\textwidth]{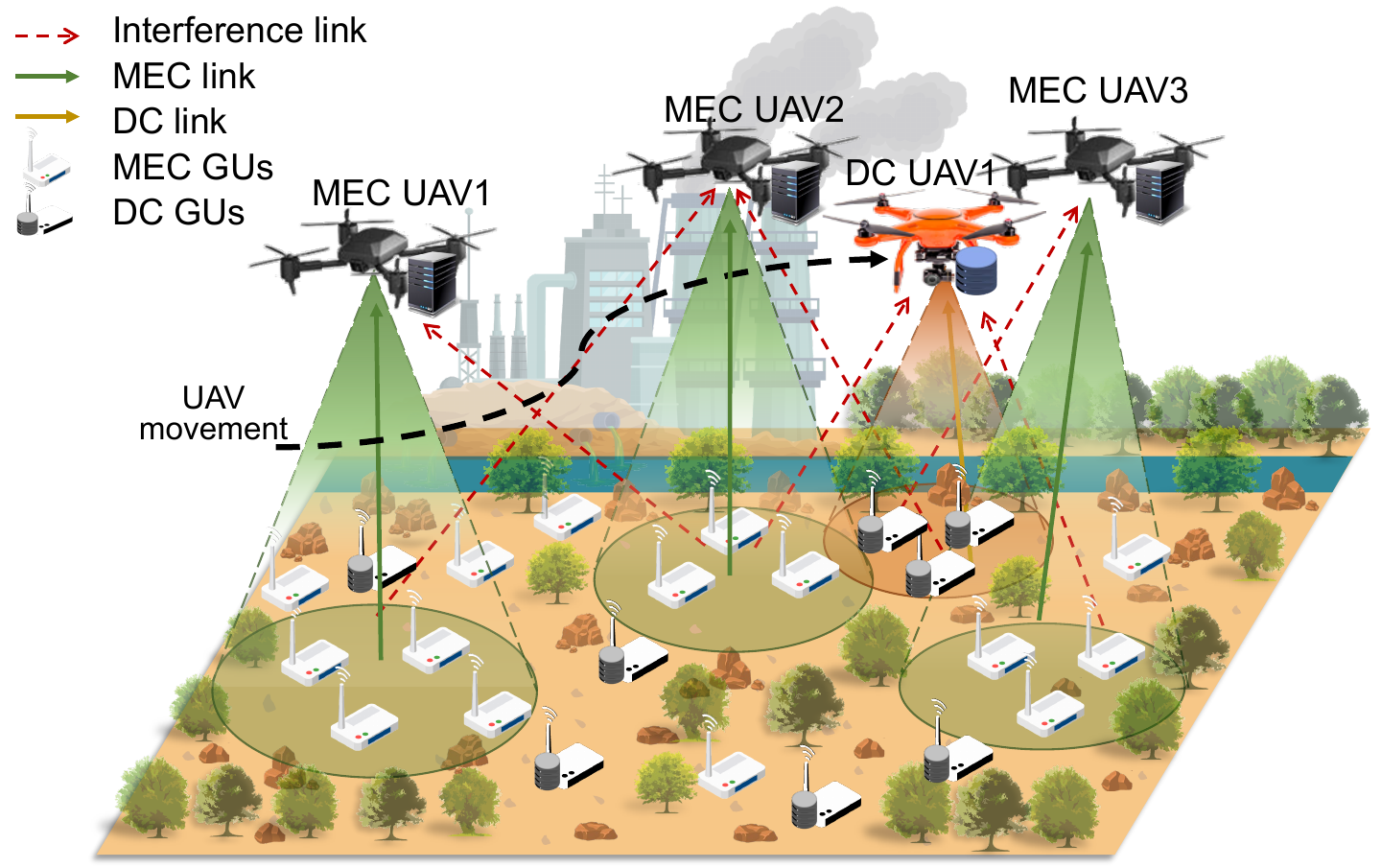}
\caption{UAV-assisted joint MEC-DC system.}
\label{fig:network_model}
\end{figure}
\par The UAV-assisted joint MEC-DC system under consideration is shown in Fig. \ref{fig:network_model}. In this system, there are massive GUs distributed in a monitored area, and these GUs focus on different functions. For example, some users may need to perform delay-sensitive computation-intensive tasks such as face recognition, image processing, and augmented reality, while another part of users need to perform time-insensitive DC tasks such as data backup, environmental monitoring, and log maintenance. Therefore, due to the different types of tasks performed, these users are divided into MEC-users denoted as $m\in \mathcal{G}^{MEC} = \left \{ 1, \dots, M \right \}$, and DC-users denoted as $n\in \mathcal{G}^{DC} = \left \{ 1, \dots, N \right \}$. These MEC-users and DC-users intermittently produce random computation-intensive tasks and time-insensitive DC tasks, respectively, and they are uniformly represented as $g\in \mathcal{G}=\mathcal{G}^{MEC} \cup \mathcal{G}^{DC}$. Since the area is remote, there is no base station can directly provide services for GUs. Moreover, due to the restricted computing and storage capabilities of GUs, processing of tasks need to be transferred to a nearby edge server or data center. Therefore, there are several UAVs provide computing or data collecting service for GUs. However, MEC requires a nearby server for stable service to meet task delay requirements, reducing the need for UAVs to move in a large area, while DC needs longer UAV flights for comprehensive data collection coverage. Thus, UAVs are divided into the MEC-UAVs denoted as $i\in \mathcal{U}^{MEC} = \left \{ 1,\dots , N_U \right \}$ and a DC-UAV $u_d$ to enhance the overall efficiency of the system. Following the completion of task offloading and edge computing, the data is returned to the GU, and the stored data collected by the DC-UAV is transmitted to a nearby base station or data center. All the UAVs are represented by a set which is $u\in \mathcal{U}=\left \{ 1, \dots, N_U, u_d \right \}$, where $u_d = N_U+1$ and $\mathcal{U}=\mathcal{U}^{MEC} \cup \{u_d\}$. 

\par The aforementioned system operates over a continuous time period, where the time horizon is divided into $T$ segments with the same duration $\tau$, which is indexed as a set $t \in \mathcal{T} = \left \{1,\dots ,T \right \}$. At each time step, the MEC-UAVs search GUs with computation requirements and provide services, while the DC-UAV approaches GUs to collect data. It is worth noting that it is generally difficult to facilitate one-to-many communication protocols due to the limited transmit power and computing capabilities of GUs. Meanwhile, to avoid further interference, we consider that a GU can only be served by one UAV while a UAV can serve multiple GUs at the same time step \cite{Chen2023}. Moreover, in the process of UAV service, since the considered UAVs use the same frequency band to communicate with the GUs, there exists co-channel interference during the communication. In addition, each UAV is equipped with an omni-directional antenna for communication with GUs and uses orthogonal frequency division multiple access (OFDMA) protocol to avoid interference among the GUs it serves \cite{zheng2022}. In this case, the uplink interference of GUs to other UAVs is not negligible and significant.

\par Without loss of generality, a 3D Cartesian coordinate system is adopted in this work, and we consider that the UAVs have a fixed altitude $H$. Therefore, the coordinates of the UAV $u$ and the GU $g$ are denoted as $V_u(t) = (x_u(t), y_u(t),H)$ and $V_g(t) = (x_g(t),y_g(t),0)$, respectively. Furthermore, we denote the distance of movement as $m(t)$ and the angle of deviation as $\alpha(t)$. Moreover, the location $V_u(t)$ can be update by $x_u(t+1) = x_u(t) + m_u(t)\cos \alpha_u(t)$ and $y_u(t+1) = y_u(t) + m_u(t)\sin \alpha_u(t)$. In the following sections, the specific UAV communication, energy consumption, MEC, and DC models are presented.

\subsection{UAV Communication Model}

\par Prior to the MEC and DC, it is imperative to establish communication links. Since the UAVs are usually maintained at a relatively high altitude, they can establish line-of-sight (LoS) channels with the GUs. Thus, the probabilistic LoS model is adopted in this work, which is given by $P_{u,g}^{LoS}(t) = 1/(1+\lambda_1 \exp\{-\lambda_2(\theta_{u,g}(t) - \lambda_1)\})$, where $\theta_{u,g}(t) = (180/\pi)\arctan(H/d_{u,g}(t))$ corresponds to the elevation angle between UAV $u$ and GU $g$, and the distance between UAV $u$ and GU $g$ is denoted as $d_{u,g}(t) = \left \| V_u(t) - V_g(t) \right \|$, and $\lambda_1$ as well as $\lambda_2$ are constant values associated with the environment \cite{Chen2017}. Moreover, the probability of non-LoS (NLoS) link between UAV $u$ and GU $g$ at the time step $t$ is given by $P_{u,g}^{NLoS}(t) = 1 - P_{u,g}^{LoS}(t)$.

\par Therefore, denote the free space path loss between GU $g$ and UAV $u$ as $L_{u,g}(t) = 20\log d_{u,g}(t)+20\log f_c+20\log(\frac{4\pi}{c})$, the average path loss between UAV $u$ and the GU $g$ at time step $t$ can be expressed as follows:
\begin{equation}
PL_{u,g}(t) = L_{u,g}(t) + P_{u,g}^{LoS}(t)\eta_{LoS} + P_{u,g}^{NLoS}(t)\eta_{NLoS},
\end{equation}
where $f_c$ and $c$ denote the carrier frequency and the velocity of light, respectively. Moreover, $\eta_{LoS}$ and $\eta_{NLoS}$ correspond to the excessive path losses for LoS and NLoS links, respectively.

\par At each time step, each UAV moves a relatively short distance with respect to the size of the considered area. Therefore, the channel gain between the UAVs and the GUs during the movement of the UAV is a quasi-constant and is calculated based on the updated position of the UAV at each time step. Specifically, the channel gain is given by
\begin{equation}\label{eq:gain}
\begin{aligned}
h_{u,g}(t) &= 10^{-\frac{PL_{u,g}(t)}{10}}\\
           &= \frac{10^{-\frac{(\eta_{LoS}-\eta_{NLoS})P_{u,g}^{LoS}(t)+\eta_{NLoS}}{10}}}{\left (\left \|V_u(t)-V_g(t)\right \| \right )^2\left (\frac{4\pi f_c}{c}\right )^2}.
\end{aligned}           
\end{equation}

\par Denote $p_g(t)$ as the transmit power used by GU $g$ at time step $t$. Then, we define an indicator variable $X_{u,g}(t)$ related to user association to indicate whether GU $g$ is associated with UAV $u$ at time step $t$. Specifically, if GU $g$ associates with UAV $u$, $X_{u,g}(t)=1$; Otherwise, $X_{u,g}(t)=0$. Thus, the data transmission rate of GU $g$ associated with UAV $u$ is given by
\begin{equation}\label{eq:rate}
R_{u,g}(t)=\frac{W}{s_u(t)}\log_2\left (1+\frac{p_g(t)h_{u,g}(t)}{I_m+n_0\frac{W}{s_u(t)}}\right ),
\end{equation}
where $I_m={\textstyle \sum_{j=1,j\ne u}^{N_U}}{\textstyle \sum_{l=1,l\ne g}^{N_g}}X_{j,l}(t)p_{j,l}(t)h_{j,l}(t)$ is the co-channel interference from all other GUs associated with other UAVs at time step $t$ and $N_g$ represents the total number of GUs, $s_u(t)=\sum_{g\in\mathcal{G}}X_{u,g}(t)$ is the total number of GUs that are associated with UAV $u$ at time step $t$, $n_0$ is the noise power spectral density and $W$ is the total bandwidth of each UAV.

\subsection{UAV Energy Consumption Model}
\par The management of the energy consumption of UAVs is necessary to ensure continuous communication and computing. Specifically, this work considers the energy consumption during UAV movement and the computation energy consumption associated with the MEC on UAVs. 

\par For a rotary-wing UAV, we denote $E_u^{move}(t)$ as the propulsion power consumption of UAV $u$ hovering and flying in a two-dimensional (2D) plane at time step $t$, and the detailed propulsion power consumption model can refer to~\cite{Yu2021}. 

\par For the energy of computation, the MEC-UAVs receive tasks from multiple GUs and perform computation during the time step $t$, and the energy consumption of computing on MEC-UAV $i$ is given by~\cite{Wang2016}
\begin{equation}
    E_i^c(t) = \kappa_i \omega_i^2 C_i l_{m,f}(t),
\end{equation}
where $\kappa_i$ represents the effective switched capacitance coefficient with respect to the CPU architecture on MEC-UAV $i$, $C_i$ represents the computation intensity on MEC-UAV $i$ (cycles/bit), $\omega_i$ is the CPU operating frequency on MEC-UAV $i$, and $l_{m,f}(t)$ is the data size of task $f_m(t)$ of MEC-GU $m$ at time step $t$.

\par To sum up, the energy consumption of the UAVs in each part discussed above is independent of each other. Therefore, the overall energy consumption of UAV $u$ at time step $t$ is defined as follows:
\begin{equation}
E_u^{total}(t)=\begin{cases}
  E_u^{move}(t)+E_u^c(t),  & \text{$u\in\mathcal{U}^{MEC}$,} \\
  E_u^{move}(t),           & \text{$u=u_{dc}$.}
\end{cases}
\end{equation}

\subsection{MEC Model}

\par Generally, GUs such as IoT devices have limited computing resources. Thus, we consider offloading all GU tasks to UAVs and returning the results to GUs after processing on UAVs. Since the task (e.g., face recognition and image processing) output is much smaller than the size of the offloaded task data, the process of returning result data to the GUs is ignored here, as in \cite{Chen2023} and \cite{seid2021}. Moreover, denote all the tasks as $f_g(t)\in \mathcal{F} = \{ 1,\dots, F \}$, we consider a deadline mechanism and the maximum tolerance time limit $t_{m,f}^{max}$ to ensure the completion rate and the completion quality of MEC, respectively. Specifically, each MEC task $f_m(t)$ has a deadline, denoted as $D_{m,f}$, and the $t^{max}_{m,f}$ is the maximum tolerance time for task $f_{m}(t)$ to be timed from the start of transmission. If the duration that task $f_m(t)$ exists after its generation exceeds $D_{m,f}$, the task becomes invalid, and task $f_m(t)$ is marked as incomplete. As such, a tuple $\{ b_{m,f}, l_{m,f}, t_{m,f}^{max}, D_{m,f}\}$ is used to characterize the MEC task $f_{m}(t)$, where $b_{m,f}$, $l_{m,f}$, $t_{m,f}^{max}$, and $D_{m,f}$ correspond to the completion status, the number of data bits, the maximum tolerable time limit, and the deadline of the task $f_{m}(t)$, respectively. Subsequently, the task generation model, transmission latency, and computation latency are as follows:

\subsubsection{Task Generation Model}

\par In real-world scenarios, the task and data generation of IoT devices within a period is usually unpredictable. Therefore, we consider an intermittent task generation model to simulate this dynamic process. Specifically, task $f_g(t)$ is definitely generated within a fixed number of time steps while the exact time step at which it is generated is random. In other words, the probability of task generation will increase according to the time variation. Since in the real world, the task requirements are dynamic rather than static, this intermittent task generation can simulate the stochastic and dynamic nature of task generation by modeling the dynamic update of real-world MEC tasks or data generation of IoT devices~\cite{Gu2021}. By using this method, the probability of task $f_g(t)$ generated by GU $g$ at time step $t$ can be given by

\begin{equation}
    P_{g,f} =\delta_g(t-\eta_g)
\end{equation}
where $\eta_g$ is the last time step of task generation of GU $g$ and the $\delta_g$ is the task density coefficient with the value interval of $(0,1)$, which reflects the speed of task generation.

\subsubsection{Transmission Latency from GUs to UAVs}

\par As aforementioned, the MEC-users need to offload computationally intensive tasks to the MEC-UAVs before performing edge computing. Thus, the data transmission latency is given by
\begin{equation}\label{eq:trans_latency}
T_{i,m}^f(t) = \frac{X_{i,m}(t)b_{m,f}(t)l_{m,f}(t)}{R_{i,m}(t)},
\end{equation}
where $b_{m,f}(t)$ denotes the task completion status by using 0 or 1 to indicate whether the task is completed or not, $l_{m,f}(t)$ represents the number of data bits of task $f_m(t)$ at time step $t$. Specifically, if the transmission of task $f_m(t)$ can be completed in this time slot, $l_{m,f}$ will remain unchanged, otherwise $l_{m,f}$ will be updated to the remaining data amount of task $f_g(t)$ in the next time step. Moreover, $R_{i,m}(t)$ represents the data transmission rate from MEC-GU $m$ to MEC-UAV $i$ at time step~$t$.

\subsubsection{Computation Latency on UAVs}

\par We consider that each GU has limited computational capabilities, and thus it is unable to perform local computing tasks. As a result, computation tasks of the GUs can be offloaded to the MEC-UAVs for edge computing. The computing latency of task $f_m(t)$ for MEC-GU $m$ on MEC-UAV $i$ is given by
\begin{equation}
T_{i,m,f}^c(t)=\frac{C_iX_{i,m}(t)b_{m,f}(t)l_{m,f}(t)}{\omega_u}.
\end{equation}

\par Thereafter, the total latency of MEC-UAV $i$ at time step $t$ is given by

\begin{equation}
T_i(t)=\sum_{g\in\mathcal{G}^{MEC}}\sum_{f\in\mathcal{F}}T_{i,m}^f(t)+\sum_{g\in\mathcal{G}^{MEC}}\sum_{f\in\mathcal{F}}T_{i,m,f}^c(t).
\end{equation}
\par Since the computation latency is related to the number of data bits of the tasks as well as the computing capacity of UAVs, only the MEC transmission latency is considered to be optimized, which will be analyzed in Section~\ref{sec:problem formulation}.

\subsubsection{MEC Task Completion Rate}

\par In the joint system, some tasks may not be completed in time due to severe interference and the limitation of computing and communication capabilities of UAVs. Therefore, task completion status is an important indicator in the MEC subsystem, and the MEC task completion rate $\mathcal{C}^{MEC}$ is defined as follows:
\begin{equation}
    \mathcal{C}^{MEC}=\frac{\sum_{i \in \mathcal{U}^{MEC}}N^f_{i}}{\sum_{m \in \mathcal{G}^{MEC}}N_m^f}\times100\%,
\end{equation}
where $N^f_{i}$ denotes the total number of tasks completed by the UAV~$i$, $N^f_{m}$ is the number of total tasks generated by the MEC-user~$m$.

\subsection{DC Model}
\par As aforementioned, a UAV collects data generated intermittently by the GUs and then transmits it to a nearby base station. For simplicity, the transmission to a nearby base station is ignored, like \cite{Du2024} and \cite{Dandapat2024}. Denote $u_{dc}$ is the DC-UAV, and the amount of collected data at time step $t$ is given by 
\begin{equation}
D(t)=\sum\nolimits_{n\in\mathcal{G}^{DC}}\tau X_{u_{dc},n} R_{u_{dc},n}(t).
\end{equation}

\par Similar to the MEC subsystem, we define the DC rate as the ratio of the amount of data collected by UAVs to the total amount of data generated by all DC-users over a period of time, which is given by
\begin{equation}
    \mathcal{C}^{DC}=\frac{\sum_{t=0}^TD(t)}{\sum_{n\in\mathcal{G}^{DC}}\sum_{t=0}^Tl_n(t)}\times 100\%,
\end{equation}
where $l_n(t)$ is the data volume generated by the DC-GU $n$ at the time step $t$.

\par During the DC process, the DC-UAV needs to approach DC-users for higher transmission rates, while this may cause greater interference with the communication of other MEC-UAVs. According to Eq.~\eqref{eq:rate}, it can be analyzed that the communication rates from GUs to UAVs are affected by the transmit power of the GUs and the location of UAVs and GUs. Since the GUs are stationary, the decision variables for DC include transmit power of the GUs and the movement of the UAVs. Additionally, since the amount of data stored by DC-GUs is different at the same time step and is subject to different levels of interference from the nearby MEC-GUs, user association is also a key decision variable for DC. Ultimately, the primary goal of DC is to collect as much data as possible. This goal is a long-term optimization problem, which is affected by various factors, and these characteristics will be discussed in detail in the following section.


\section{Problem Formulation }\label{sec:problem formulation}
\par In the considered UAV-assisted joint MEC-DC system, GUs intermittently generate computation-intensive tasks and freshness-insensitive data, and the MEC-UAVs and the DC-UAV perform MEC and DC, respectively. The main goal of the system is to reduce the system latency of MEC while maximizing the volume of collected data.

\par However, the communication between the UAVs and the GUs can cause co-channel interference, affecting the QoS of other GUs. Specifically, the MEC delay and the amount of collected data are both determined by the transmission rate, which is mainly affected by user transmit power, UAV movement, and user association according to the Eqs. \eqref{eq:gain}-\eqref{eq:trans_latency}. Although optimizing these variables can improve the transmission rates of some GUs, this will also increase the interference, resulting in additional total system delay and a reduction in the amount of collected data. Therefore, the user transmit power, UAV movement, and user association are interdependent and coupled decision variables. Moreover, due to the interaction among these decision variables, the system needs to consider these decision variables comprehensively.

\par The decision variables consist of three parts of variables, which are the movement of UAVs, user transmit power, and user association, respectively. We define these decision variables as follows. \textit{(i)} The positions of UAVs are represented by the matrix $\boldsymbol{V}=\{V_u(t)| u\in\mathcal{U}, t\in \mathcal{T}\}$ and it is continuous. \textit{(ii)} The transmit power value of each GU is represented by a continuous variable, and the transmit power of all GUs is represented by the vector $\boldsymbol{p}=\{p_g(t)| g\in\mathcal{G}, t\in\mathcal{T}\}$. \textit{(iii)} The association relationship between the UAVs and the GUs is represented by a discrete matrix $\boldsymbol{X}=\{X_{u,g}(t)| u\in\mathcal{U}, g\in\mathcal{G}, t\in\mathcal{T}\}$. Consequently, the formulation of optimization objectives are as follows.

\par \textbf{\textit{Optimization Objective 1:}} The first objective is to reduce the total system latency of MEC because the considered MEC subsystem has high latency requirements. Thus, the first objective function is given by
\begin{equation}
    f_1 \left ( \boldsymbol{V}, \boldsymbol{p}, \boldsymbol{X} \right ) = \sum\nolimits_{t=1}^{T}\sum\nolimits_{i\in\mathcal{U}^{MEC}}T_i(t).
\end{equation}

\par \textbf{\textit{Optimization Objective 2:}} The second objective is to increase the amount of collected data by the DC-UAV as it determines the maximum DC capability of the considered DC subsystem, i.e.,
\begin{equation}
    f_2 \left ( \boldsymbol{V}, \boldsymbol{p}, \boldsymbol{X} \right ) = \sum\nolimits_{t=1}^{T} D(t),
\end{equation}

\par According to the two objectives above, the considered problem is formulated as follows:
\begin{subequations}
\begin{align}
    \mathcal{P}: \min_{\{\boldsymbol{V}, \boldsymbol{p}, \boldsymbol{X}\}} &Q= \{f_1,-f_2\}, \\
    \text{s.t.}~~&T_{i,m}^f(t) \le T_{m,f}^{max}, \forall i \in \mathcal{U}^{MEC}, \forall m \in \mathcal{G}^{MEC},\nonumber\\ 
    & \forall t\in\mathcal{T}, \label{eq:subb}\\
    &p_g(t) \le p_g^{max},  \forall g \in \mathcal{G}, \forall t\in\mathcal{T}, \label{eq:subc}\\
    &\sum_{t \in \mathcal{T}}E_u^{total}(t) \le E_u^{max}, \forall u \in \mathcal{U}, \forall t\in\mathcal{T}, \label{eq:subd}\\
    &X_{u,g}(t) \in \{0,1\}, \forall u \in \mathcal{U}, \forall g \in \mathcal{G}, \forall t\in\mathcal{T}, \label{eq:sube}\\
    &\sum\nolimits_{u\in\mathcal{U}}X_{u,g}(t) = 1, \forall g \in \mathcal{G}, \forall t\in\mathcal{T}, \label{eq:subf}\\
    &\sum_{g\in\mathcal{G}}X_{u,g}(t) \le N_u^{max}, \forall u \in \mathcal{U}, \forall t\in\mathcal{T}, \label{eq:subg}\\
    &x_{min} \le x_u(t) \le x_{max}, \forall u \in \mathcal{U}, \forall t\in\mathcal{T}, \label{eq:subh}\\
    &y_{min} \le y_u(t) \le y_{max}, \forall u \in \mathcal{U}, \forall t\in\mathcal{T}, \label{eq:subi}\\
    &0 \le m_u(t) \le m_u^{max}, \forall u \in \mathcal{U}, \forall t\in\mathcal{T}, \label{eq:subj}\\
    &\left \| V_a(t)-V_b(t) \right \| \ge d_{min}, \forall a, b \in \mathcal{U}, a \ne b,\nonumber\\
    & \forall t\in\mathcal{T}, \label{eq:subk}
\end{align}
\end{subequations}
where \eqref{eq:subb}, \eqref{eq:subc}, and \eqref{eq:subd} provide constraints on latency, transmit power of GUs, and UAV energy consumption, respectively. Moreover, \eqref{eq:sube}, \eqref{eq:subf}, and \eqref{eq:subg} provide value ranges of associated indicator variables and constraints on GU associations, respectively. In addition, \eqref{eq:subh} and \eqref{eq:subi} limit the moving range of UAVs in X and Y directions, respectively, \eqref{eq:subj} restrict the maximum moving distance of UAV within a time step, and \eqref{eq:subk} is to avoid collisions among UAVs.

\par The problem $\mathcal{P}$ has the following properties that make it challenging to solve. \textit{First}, it is a mixed-integer non-convex programming problem. Specifically, the problem $\mathcal{P}$ has one integer decision variable $\boldsymbol{X}$ and two continuous decision variables $\boldsymbol{V}$ and $\boldsymbol{p}$. Moreover, the problem exhibits non-convex properties since the constraints~\eqref{eq:subb} and \eqref{eq:subc} are non-convex~\cite{Chen2023}. \textit{Second}, the problem $\mathcal{P}$ is a long-term optimization problem that aims to maximize the amount of collected data and minimize the total MEC latency over a period. This long-term issue is affected by the dynamics of stochastic generation of tasks and the current task status of users, and whenever the location of a UAV changes, the communication rates between all other UAVs and GUs are altered due to interference. \textit{Finally}, the two optimization objectives are contradictive that are difficult to balance. For example, when MEC-UAVs are associated with a high density of GUs or when the transmit power of MEC-users is high, the transmission rate of the DC-UAV will be affected because of the strong interference from the MEC subsystem. Conversely, when the DC-UAV chooses an optimal DC route, it can severely interfere with the communication of MEC-UAVs. Furthermore, as mentioned above, the three parts of decision variables are also interdependent and coupled with each other.


\par In this case, DRL can be a viable online algorithm to the problem $\mathcal{P}$ with dynamic adaptability, which is suitable for dynamic decision-making in long-term optimization problems \cite{zhao2023deep}. In the following section, the proposed solution is detailed.


\section{The Proposed DRL-based Approach}\label{sec:proposed algorithm}

\par
\begin{figure*}[htbp]
\centering
\includegraphics[width=\textwidth]{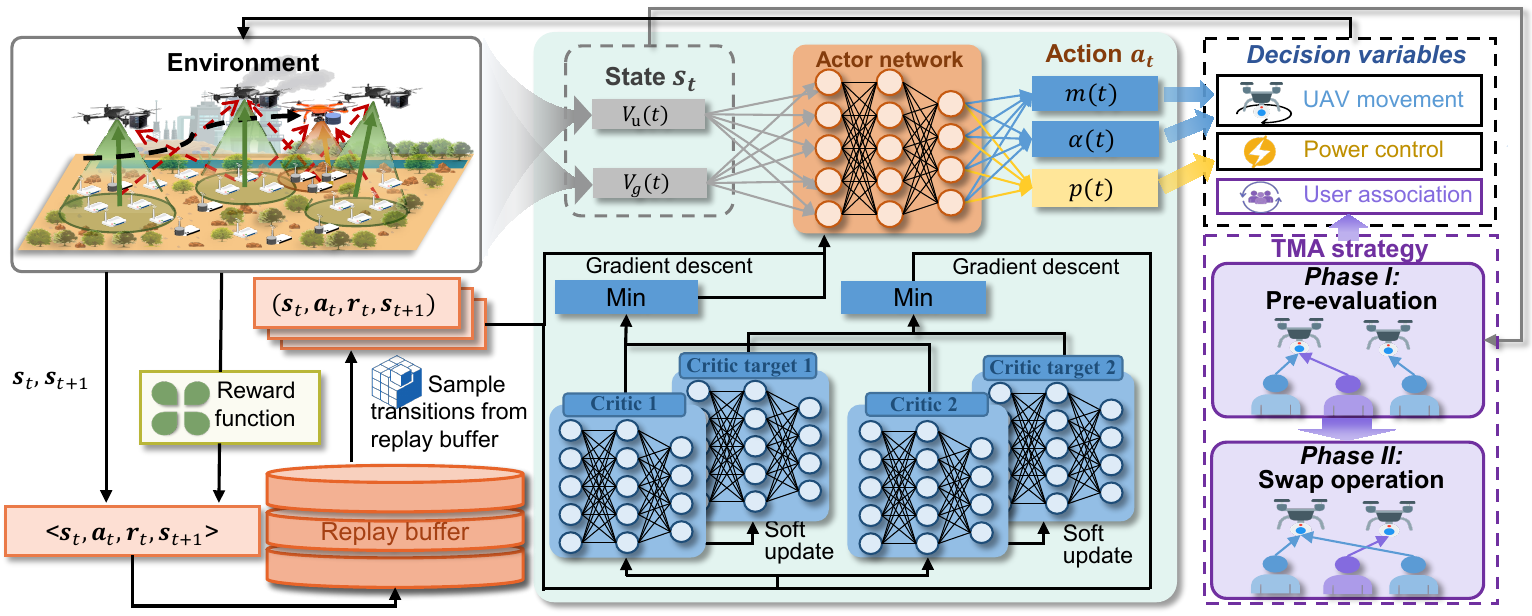}
\caption{Schematic of the proposed SAC-TMA algorithm.}
\label{fig:schematic}
\end{figure*}

 In this section, we propose the SAC-TMA to solve the formulated optimization problem, and the schematic of the SAC-TMA is shown in Fig.~\ref{fig:schematic}.

\subsection{MDP Simplification and Formulation}

\par At each time step, we aim to optimize the movement of the UAVs, the transmit power of the GUs, and the user association to minimize the MEC system sum latency while maximizing the amount of collected data. This process can be modeled as an MDP, which consists of a five-element tuple $<\mathcal{S}, \mathcal{A}, \mathcal{R}, \boldsymbol{P}, \gamma>$, where $\mathcal{S}$, $\mathcal{A}$, $\mathcal{R}$, $\boldsymbol{P}$, and $\gamma$ correspond to state space, action space, reward, probability of state transition, and discount factor, respectively.

\par In general, the action space should contain all decision variables (such as $\boldsymbol{V}$, $\boldsymbol{p}$, and $\boldsymbol{X}$) in an optimization problem when this problem is represented as an MDP. However, the optimization of the aforementioned decision variables usually results in a large action space, and there are both discrete and continuous variables. Specifically, the user association is a discrete variable with a dimension of $M\times N$. Moreover, the solution space of user association will increase exponentially in number as the number of GUs and UAVs rises, which will difficult for DRL to train and converge \cite{Wang2023}. In this case, we aim to divide the decision variables into two parts, one part is user association, which is optimized by using a separate strategy as a substitute for action, and the other part is UAV movement and user transmit power, which is optimized as the actions of the MDP. The key challenge of this task is maintaining the synergy between the optimization of user association and other decision variables, while ensuring that the optimization process for user association is feasible in computational complexity and stable.

\subsubsection{Action Reduction}\label{sec:matching}
\par To ensure collaborative optimization with DRL algorithms, a low-complexity strategy is required. Compared with traditional solutions such as exhaustive search and random methods, the matching method usually has both low complexity and effectiveness in user association problems \cite{Deng2022}. This prompts us to propose a matching-based association strategy to optimize user association decision variables. As aforementioned, each GU can be associated with a UAV, while each UAV has the capacity to serve multiple GUs. This process can be formulated as a one-to-many matching problem and can be modeled by using a two-sided matching game as follows:
\begin{definition}
    \textit{The one-to-many matching game under consideration is comprised of two groups of players, $\mathcal{G}$ and $\mathcal{U}$, and a set of association pairs is a matching denoted by $X$ when it satisfies:}
\end{definition}
\begin{enumerate}
\item $X(i) \in \mathcal{G}^{MEC}$, $X(u_{dc}) \in \mathcal{G}^{DC}$, $X(m) \in \mathcal{U}^{MEC}$, and $X(n)=u_{dc}$,
\item $\left |X(g)\right | = 1$ and $\left |X(u)\right | \leq N_u^{max}$,
\item $g \in X(u) \Leftrightarrow X(g)=u$,
\end{enumerate}
where condition 1) indicate that MEC-users are exclusively associated with MEC-UAVs, while DC-users are confined to be associated with the DC-UAV. Condition 2) further represent the constraints on the association of GUs and UAVs, specifying that each GU $g$ is restricted to associate with a single UAV, and conversely, each UAV is limited to be associated with a maximum of $N_u^{max}$ GUs. The mutual association between a GU $g$ and a UAV $u$ is defined in condition 3), asserting that if GU $g$ is associated with UAV $u$, then UAV $u$ is correspondingly associated with GU $g$.

\par Therefore, the indicator of user association at time step $t$ can be specified as $X_{u,g}(t)$ from matching $X$, which uses 1 and 0 to denote GU $g$ is associated with UAV $u$ or not.

\par Since both MEC latency and amount of collected data depend on the communication rate from GUs to UAVs, we select the system sum rate as the utility of the aforementioned matching model, with all players pursuing a common goal to maximize the system sum rate, the utility is defined by
\begin{subequations}
\label{eq:utility}
\begin{align}
    U(X)&=\sum\nolimits_{u\in \mathcal{U}}\sum\nolimits_{g\in \mathcal{G}}X_{u,g}(t)R_{u,g}(t),\\
    \text{s.t.}~~&R_{u,g} \ge R_{M_{th}}, \forall u \in \mathcal{U}^{MEC}, \forall g \in \mathcal{G}^{MEC}, \\
    &R_{u,g} \ge R_{D_{th}}, \forall u = u_{dc}, \forall g \in \mathcal{G}^{DC},
\end{align}
\end{subequations}
where $R_{M_{th}}$ and $R_{D_{th}}$ denote the threshold rates for MEC and DC, respectively.

\par Due to the fact that the communication rate from each GU to UAVs is affected by the interference of other GUs, when the association of one GU changes, the association preferences of other GUs will also change accordingly. This interdependency among GUs is known as externality \cite{bando2012,Zhao2017} and the one-to-many matching problem with externalities can be solved by swap matching~\cite{BodineBaron2011}. However, the random initialization in the conventional swap method may lead to a slow convergence of the swap process and a tendency to be stuck in the local optimum. Therefore, we propose a two-phase swap-based association strategy that enhances the initial matching. In what follows, the two phases of the proposed strategy are given in detail.

\par \textbf{Phase I: Gale-Sharpley (GS)-based pre-evaluation.} Since the swap algorithm can continue to optimize at any time, the initial matching can be obtained by using the method based on GS, which is widely used in matching problems~\cite{Huo2024}. Specifically, to maximize the utility, an evaluation method based on the communication rates can be used to obtain an initial matching. However, the calculation of communication rates requires the association between other GUs. Therefore, we consider two pre-evaluation methods to determine the initial matching in two steps, which are distance-based evaluation and rate-based evaluation, respectively.

\begin{itemize}
    \item For distance-based evaluation, GUs first choose the nearest UAV to make requests. If the UAV has more GUs than its service capacity, it will sort all served GUs in ascending order of distance and reject those GUs that are beyond capacity. The rejected GUs will then turn to the next nearest UAV to send requests. The distance-based evaluation algorithm is shown in Algorithm \ref{alg:distance based}.
    \item For rate-based evaluation, the initial matching is first obtained base on the distance-based evaluation method. Subsequently, for each GU, it calculates the communication rate with each UAV, and selects the UAV with the highest rate to send a request. If the UAV has more GUs than its service capacity, it will sort all served GUs in ascending order of rate and reject those GUs that are beyond capacity. The rejected GUs will then turn to the UAV that has the next highest rate and make requests. In addition, to ensure the communication quality of MEC and DC, the association is only allowed when the achievable communication rate between the GU and the UAV is not less than the preset threshold rate. The rate-based evaluation algorithm is shown in Algorithm \ref{alg:rate based}.
\end{itemize}

\begin{algorithm}[tbp]
    \small
    \caption{Distance-based evaluation algorithm}
    \label{alg:distance based}
    \textbf{Initialize:} Create an available UAV list $\mathcal{I}_m$ and set variable $\mathrm{CONS}_m \leftarrow \mathrm{False}$ for $m \in \mathcal{G}^{MEC}$\;
    \emph{// MEC-user association}\\
    \While{$\exists m, \mathrm{CONS}_m = \mathrm{False}$ and $\mathcal{I}_m \ne \emptyset$}{
        Choose the nearest MEC-UAV $i \in \mathcal{I}_m$\;
        $X_{i,m}(t) \leftarrow 1$\;
        $\mathrm{CONS}_m \leftarrow \mathrm{True}$\;
        \If{$\sum_{m \in \mathcal{G}^{MEC}}X_{i,m}(t) > N_u^{max}$}{
        Find the farthest GU $m_f$\;
        $X_{i,m_f}(t) \leftarrow 0$\;
        $\mathcal{I}_{m_f} \leftarrow \mathcal{I}_{m_f} \backslash \{i\}$\;
        $\mathrm{CONS}_{m_f} \leftarrow \mathrm{False}$\;
        }
    }
    \emph{// DC-user association}\\
    \ForAll{$n \in \mathcal{G}^{DC}$}{
        \eIf{$L_n(t) < D_{min}$}{
            $X_{u_{dc},n}(t) \leftarrow 0$\;
            }{
            $X_{u_{dc},n}(t) \leftarrow 1$\;
            \If{$\sum_{n \in \mathcal{G}^{DC}}X_{u_{dc},n}(t) > N_u^{max}$}{
                Find the farthest GU $n_f$\;
                $X_{u_{dc},n_f}(t) \leftarrow 0$
            }
            }
    }
    Return $X(t)$\;
\end{algorithm}

\par \textbf{Phase II: Swap-based matching.} After the initial matching is achieved in the first phase, the swap method is used to optimize utility by switching matching pairs between GUs and UAVs to reach an optimal state, and the definition of swap matching is as follows. 
\begin{definition}
    \textit{Given a one-to-many matching $X$ with $g \in X(u)$, $g^{'} \in X(u^{'})$, the swap matching of GU $g$ and GU $g'$ is defined by $X_g^{g'}=\{X \backslash \{(u,g),(u',g')\} \cup \{(u,g'),(u',g)\}\}$.}
\end{definition}

\par Swap matching enables the exchange of associated UAVs between two different GUs without affecting the association of other GUs with UAVs. Note that the single-UAV DC scenario is a special one-to-many matching, where there exists a user pair $(n, n')$ with $n \in X(u_{dc})$ and $\left |X(n')\right |=0$. Thus, we define the swap matching in this case as $X_n^{n'}=\{X \backslash \{(u_{dc},n),(emp,n')\} \cup \{(u_{dc},n'),(emp,n)\}\}$, where $emp$ denotes a state of not being associated with any UAV. Subsequently, the stable matching is given below.
\begin{definition}
    \textit{A matching $X$ is considered stable if and only if no swap pairs exist that could improve the matching.}
\end{definition}

\par Definition 3 points out that, when a matching reaches the stable state, there exists no user pairs $(g, g')$ with $g \in X(u)$ and $g' \in X(u')$ such that $U(X_g^{g'}) >U(X)$, where $(g, g')$ is referred to as a swap-blocking pair.

\par The main steps of the two-phase swap-based association strategy is outlined in Algorithm \ref{alg:two-phase}. In \textbf{Phase I}, a preliminary stable matching is obtained based on the positions of the GUs and UAVs. Subsequently, the communication rates from GUs to UAVs are calculated according to the established matching. Finally, by assessing the utility, a new stable matching is derived. In \textbf{Phase II}, the swap algorithm is utilized to iteratively optimize the matching from the previous phase. It is worth noting that during the iterative process, each swap operation of user pairs ensures a strict improvement in the matching effect. Moreover, since the number of GUs and UAVs is finite and usually small, this ensures that the proposed TMA strategy can complete the iterations within a finite number of steps and converge to a stable matching~\cite{Wang2023,Huo2024}. As a result, the stable matching that is ultimately obtained represents the optimized user association after the optimization process.

\begin{algorithm}[tbp]
    \small
    \caption{Rate-based evaluation algorithm}
    \label{alg:rate based}
    \textbf{Initialize:} Obtain an association $X^{\circ}$ by Algorithm \ref{alg:distance based}, create an available UAV list $\mathcal{I}_m$ and set variable $\mathrm{CONS}_m \leftarrow \mathrm{False}$ for $m \in \mathcal{G}^{MEC}$\;
    \emph{// MEC-user association}\\
    \While{$\exists m, \mathrm{CONS}_m = \mathrm{False}$ and $\mathcal{I}_m \ne \emptyset$}{
        Calculate $R_{i,m}(t)$ by using Eq.~\eqref{eq:rate} and $X^{\circ}$\;
        Choose $i_h \in \mathcal{I}_m$ with the highest $R_{i_h,m}(t)$\;
        \eIf{$R_{i_h,m}(t) \ge R_{M_{th}}$}{
            $X_{i_h,m}(t) \leftarrow 1$\;       
            \If{$\sum_{m \in \mathcal{G}^{MEC}}X_{i_h,m}(t) > N_{i_h}^{max}$}{
                Find $m_l$ with the lowest $R_{i_h,m_l}$\;
                $X_{i_h,m_l}(t) \leftarrow 0$\;
                $\mathcal{I}_{m_l} \leftarrow \mathcal{I}_{m_l}\backslash \{i_h\}$\;
                $\mathrm{CONS}_{m_l} = \mathrm{False}$\;
        }
        }{
        $X_{i_h,m}(t) \leftarrow 0$\; 
        }
        $\mathrm{CONS}_m \leftarrow \mathrm{True}$\;
        }
    
    \emph{// DC-user association}\\
        \ForAll{$n \in \mathcal{G}^{DC}$}{
            \eIf{$l_n(t) < D_{min}$}{
            $X_{u_{dc},n}(t) \leftarrow 0$\;
            }{
            \eIf{$R_{u_{dc},n}(t) \ge R_{D_{th}}$}{
                $X_{u_{dc},n}(t) \leftarrow 1$\;
            \If{$\sum_{n \in \mathcal{G}^{DC}}X_{u_{dc},n}(t) > N_u^{max}$}{
                Find $n_l$ with the lowest $R_{u_{dc},n_l}$\;
                $X_{u_{dc},n_l}(t) \leftarrow 0$\;
            }
            }{
                $X_{u_{dc},n}(t) \leftarrow 0$\;
            }
            }
    }

    Return $X(t)$\;
\end{algorithm}

\begin{algorithm}
    \small
    \caption{Two-phase matching-based association (TMA)}
    \label{alg:two-phase}
    \textbf{Initialization:} Obtain the location of GUs and UAVs. Calculate the distance matrix of GUs and UAVs\;
    \textbf{Phase I:} GS-based pre-evaluation\\
    Obtain an initial association $X$ by using Algorithm~\ref{alg:rate based}\;
    Calculate Eq. \eqref{eq:utility}\;
    \textbf{Phase II:} Swap-based matching\\
    \emph{// MEC-user association}\\
    \While{Swap-blocking pairs exists}{
    Choose $m, m' \in \mathcal{G}^{MEC}$, $i = X(m)$, and $i'=X(m')$\;
    Calculate Eq. \eqref{eq:utility}\;
    \If{The user pair $(m, m')$ is a swap-blocking pair}{
    $X \leftarrow X_m^{m'}$\;
    }
    }
    \emph{// DC-user association}\\
    \While{Swap-blocking pairs exists}{
    Choose $n, n' \in \mathcal{G}^{DC}$, $j = X(n)$, $\left |X(n')\right |=0$\;
    Calculate Eq. \eqref{eq:utility}\;
    \If{The user pair $(n, n')$ is a swap-blocking pair}{
    $X \leftarrow X_n^{n'}$\;
    }
    }
    \textbf{return} The stable matching-based association~$X$.
\end{algorithm}

\par As such, after the GS-based pre-evaluation, the swap approach starts the iterative search for swap-blocking pairs, updates the association, and calculates the utility, until a stable matching is reached, thus determining the final matching, which is the optimized user association. In this case, we can use the final stable matching as the instantaneous user association, which can replace the user association action in the MDP by integrating the matching result into the MDP as part of the output action, which reduces the training difficulty of DRL algorithms and ensures stability.

\subsubsection{MDP Formulation}

\par Leveraging the above simplification, the aforementioned optimization problem can be re-formulated as an action space-reduced MDP. Given the user association, we optimize the movement of the UAVs and the transmit power of the GUs to minimize the MEC system sum latency while maximizing the amount of collected data at each time step. This process can be modeled as an MDP, which consists of a five-element tuple $<\mathcal{S}, \mathcal{A}, \mathcal{R}, \mathcal{P}, \gamma>$, where $\mathcal{S}$, $\mathcal{A}$, $\mathcal{R}$, $\mathcal{P}$, and $\gamma$ correspond to the state space, action space, reward, probability of state transition, and discount factor, respectively. The detailed key definitions of MDP for the considered problem are given as follows:

\par \textbf{State:} The state at time step $t$ is denoted by $s_t$ and it consists of six parts:
    
    \begin{itemize}
        \item The position coordinate of UAV $u$ at time step $t$ $(x_u(t), y_u(t))$: Since UAVs are assumed at fixed height, the coordinates of UAVs are only consist of $(x,y)$ in the horizontal plane.
        \item The position coordinate of GU $g$ at time step $t$ $(x_g(t), y_g(t))$.
        \item The total data length of all the unfinished tasks of MEC-GU $m$ at time step $t$ $l_{m}(t)$: If MEC-GU $m$ has no tasks to compute at time step $t$, the $l_m(t)$ equals 0.
        \item The remaining processing time of the earliest unfinished task of MEC-GU $m$ at time step $t$ $D_{m,f}(t)$: If $D_{m,f}(t)$ equals 0, the task $f_g(t)$ is marked as incomplete.
        \item The amount of data storage for the DC-GU $n$ at time step $t$ $L_n(t)$.
        \item The number of remaining time steps $T^{\circ}$.
    \end{itemize}
        
\par Therefore, state $s_t$ can be written as $s_t=[x_{1}(t), y_{1}(t), \dots, x_{u}(t), y_{u}(t), \dots, x_{N_U+1}(t), y_{N_U+1}(t), x_{1}(t)$, $y_{1}(t), \dots, x_{g}(t), y_{g}(t), \dots, x_{N_g}(t), y_{N_g}(t), l_1(t), \dots, l_m(t)$, $D_{1,f}(t), \dots, D_{m,f}(t), L_1(t), \dots, L_n(t), T^{\circ}]$, the cardinality of $s_t$ is $2 \times (N_U+1) + 2 \times N_g + m + m + n +1$, where $N_U$ and $N_g$ are the numbers of UAVs and GUs respectively.

\par \textbf{Action:} As aforementioned, the UAV movement and the user transmit power are optimized as another part to minimize the MEC system sum latency while increasing the amount of collected data. Therefore, the action $a_t$ consists of two parts: UAV movement and user transmit power control, where the UAV movement is determined by the moving distance and direction of flight (i.e., yaw angle):
        \begin{itemize}
            \item $m_{u}(t)$: The moving distance of UAV $u$ at time step $t$.
            \item $\alpha_{u}(t)$: The flying direction of UAV $u$ at time step $t$.
            \item $p_{g}(t)$: Transmit power value of GU $g$ at time step $t$.
        \end{itemize}
        
\par Formally, action $a_t$ can be written as $a_t=[m_{1}(t), \dots, m_{u}(t), \dots, m_{N_U+1}(t), \alpha_{1}(t), \dots, \alpha_{u}(t), \dots$, $\alpha_{N_U+1}(t), p_{1}(t), \dots, p_{g}(t), \dots, p_{N_g}(t)]$. The cardinality of $a_t$ is $2 \times (N_U+1) + N_g$, where $N_U$ and $N_g$ are the number of UAVs and GUs, respectively. Since both the distance and direction of UAV movement and the power value of GUs are continuous, the $a_t$ is also continuous.

\par \textbf{Reward:} The primary goal of the considered optimization problem is to minimize the MEC system sum latency and maximize the amount of collected data. Thus, the reward is defined as the sum of the latency reward, the DC reward, and the penalty reward, which are defined as follows.

    
\par \textit{Latency reward:} The latency reward at time step $t$ is related to the MEC task offloading time and its maximum latency limit, which is given by
\begin{equation}\label{eq:latency reward}
    r_l(t) = \sum\nolimits_{u \in \mathcal{U}^{MEC}} (t_{m,f}^{max} - T_i^f(t)),
\end{equation}
where $T_i^f(t)$ is the latency of MEC task offloading and executing at this time step. The latency reward $r_l(t)$ means that when the MEC latency exceeds the latency limit, the agent receives a negative reward.

\par \textit{DC reward:} The reward of DC at time step $t$ is related to the size of the data stored in the GU associated with the DC-UAV. Due to the limited storage of GUs, once the generated data reaches the upper limit, the older data will be discarded. To avoid data loss, we apply a reward decay mechanism. The DC reward is defined as follows:
    \begin{equation}\label{eq:dc reward}
            r_d = \sum\nolimits_{n \in \mathcal{G}^{DC}} \sigma X_{u_{dc},n}(t)L_n(t),
    \end{equation}
    where $L_n(t)$ is the amount of stored data for GU $n$ at time step $t$, $L_n^{max}$ is the data storage limit of the GU $n$, and $\sigma$ is a discount coefficient, which is equal to 0.5 if $L_n(t)$ reaches the data storage limit $L_n^{max}$, and equal to 1 otherwise.
    
\par \textit{Penalty reward:} We define some rewards related to punishment to address situations where the constraints are not satisfied, which is defined as follows.
    \begin{itemize}
        \item UAVs collision: To avoid the collision between UAVs, we give a negative reward when there are two UAVs fly too close to each other. The penalty reward is given by
        \begin{equation}
            r_p = -\varrho,
        \end{equation}
        where $\varrho$ is a positive constant number.
        
        \item UAVs fly outside the area boundary: Due to the interference effect in the considered network, some UAVs may attempt to move to a farther position to avoid this effect, which leads to additional energy consumption and lower MEC efficiency, as all GUs are within area. To mitigate this issue, we set a penalty reward as the ratio of the out-of-bounds distance of UAV $u$ to the maximum moving distance of UAV $u$, which can punish the agent when UAVs go out of bounds while avoiding the accumulated penalty reward value to be too large leading to unstable training. The penalty reward is designed as follows:
        \begin{equation}
            r_p = r_p +\frac{\sqrt{(B_u^x(t))^2+(B_u^y(t))^2} }{\rho \cdot m_u^{max}}, 
        \end{equation}
        where $\rho$ represents the penalty factor when UAVs move beyond the border, and $B_u^x(t)$ and $B_u^x(t)$ denote the distances that UAV $u$ cross the $X$-axis and $Y$-axis boundaries at time step $t$, respectively. Specifically, the $B_u^x(t)$ and $B_u^y(t)$ are given by $B_u^x(t) = \max \{\left | x_u(t) \right | , (x_{max}-x_{min})/{2} \} - (x_{max}-x_{min})/{2},$ $B_u^y(t) = \max \{\left | y_u(t) \right | ,(y_{max}-y_{min})/{2} \} - (y_{max}-y_{min})/{2}$, respectively. Additionally, the center of the considered area is set as the origin of the coordinates $(0,0)$ and when the UAV $u$ is inside the boundary, $B_u^x(t)$ and $B_u^y(t)$ are equal to 0.
        
        \item Excessive energy consumption by UAVs: Due to limited energy carried by a UAV, onboard energy should be utilized properly. In the considered scenario, multiple UAVs provide service for GUs simultaneously and thus each UAV does not need to move long distance frequently to ensure service coverage. Therefore, we impose an energy constraint to restrict energy consumption at each time step, so that the total energy consumption of UAVs does not exceed the total energy carried by them. The penalty reward is designed as follows:
        \begin{equation}\label{eq:energy penalty}
            r_p= r_p + \delta_p \sum\nolimits_{u \in \mathcal{U}}(E_u^{total}(t)-\frac{E^{max}}{T}),
        \end{equation}
        where $\delta_p$ denotes the penalty coefficient that UAVs energy consumption exceeding the limit, $E^{max}$ denotes the total energy limit for each UAV.
    \end{itemize}
    
\par As a result, the reward can be written as follows:
\begin{equation}
    r(t) = r_l(t) + r_d(t) + r_p(t).
\end{equation}

\subsection{The Proposed SAC-TMA Algorithm}

\par Since this paper investigates the problems of continuous UAV movement and user transmit power control, and the state transfer probability is unknown, we employ model-free DRL with continuous action space to address the optimization problem related to UAV movement and user transmit power.

\par SAC is a model-free, off-policy DRL method \cite{Haarnoja2018}. Since it adopts the principle of maximum entropy DRL, which not only aims to maximize the cumulative reward but also encourages exploration of the policy space, it is beneficial for accelerating policy learning and avoiding being stuck in the local optimal points. Therefore, we employ the SAC algorithm to address the optimization of UAV movement and user transmit power. 

\par The employed SAC consists of a policy network $\pi_{\phi}(a_t|s_t)$, two critic networks $Q_{\theta_1}(s_t,a_t)$, $Q_{\theta_2}(s_t,a_t)$, and two target critic networks $Q_{\overline{\theta}_1}(s_t,a_t)$, $Q_{\overline{\theta}_2}(s_t,a_t)$. In addition, the policy entropy in SAC is defined as the level of randomness of the policy, which is given by $\mathbb{E}_{a_t\sim \pi}[-\log (\pi (a_t|s_t))]$, and the aim of SAC is to increase both the cumulative reward and the expected entropy of the policy, which is defined as follows:
\begin{equation}\label{eq:entropy}
    \begin{aligned}
        \mathit{J} (\pi)= \sum_{t=1}^T\mathbb{E}_{(s_t,a_t) \sim \rho_\pi}\Big[r(s_t,a_t)-\alpha \log \big(\pi (\cdot | s_t)\big)\Big],
    \end{aligned}
\end{equation}
where $\alpha$ is the temperature parameter used to adjust the importance of entropy and control the stochasticity of policy. This mechanism of maximum entropy encourages UAVs to explore a greater variety of potential trajectories in real-time changing environments, while the structure of two critics and two target critics helps reduce estimation bias, thereby guiding the actor for more effective exploration and exploitation, enhancing the stability of learning.

\par At every regular interval during the training process, the SAC performs gradient descent to the critic networks and the policy network. Consequently, the loss function of each critic network is defined as:
\begin{equation}\label{eq:lossQ}
    \begin{aligned}
        &L_Q(\theta) = \mathbb{E}_{(s_t,a_t)\sim \mathcal{R}^{\circ} }\Big[ \frac{1}{2} \Big( Q_{\theta}(s_t,a_t)- \Big( r_t  \\
        &+\gamma \Big(\min_{j=1,2} Q_{\overline{\beta}_j}(s_{t+1},a_{t+1})-\alpha \log \pi_\phi(a_{t+1}|s_{t+1})\Big) \Big) \Big)^2 \Big],
    \end{aligned}
\end{equation}
where $\mathcal{R}^{\circ}$ is the distribution of sampled transitions. This offline sampling mechanism allows the algorithm to reuse historical data for learning, which is very useful for scenarios with high data collection costs, such as user transmit power control. The parameters of critic networks $\theta_i, i=1,2$, are updated by minimizing the $L_Q(\theta_i)$. The parameter of the policy network $\phi$ is updated by
\begin{equation}\label{eq:lossp}
    \begin{aligned}
        L_{\pi}(\phi) = &\mathbb{E}_{s\sim \mathcal{R}^{\circ}, \epsilon_t \in \mathcal{N}_g }\Big[ \alpha \log \pi_\phi \big( f_{\phi}(\epsilon_t; s_t)| s_t \big) \\
        &- \min_{j=1,2}Q_{\theta_j}\big(s_t,f_{\phi}(\epsilon_t; s_t)\big)\Big].
    \end{aligned}
\end{equation}

\par In Eq. \eqref{eq:lossp}, the reparameterization trick is used to obtain a solution of policy gradient in continuous action space, in which the policy is reformulated as $a_t=f_{\phi}(\epsilon_t;s_t)$, with $\epsilon_t$ being an independent random noise variable.

\begin{algorithm}[tb]
    \small
    \caption{SAC-based deep-reinforcement-learning with TMA strategy (SAC-TMA)}
    \label{alg:sac-tma}
    \textbf{Initialize:} The replay buffer $\boldsymbol{R}$, parameters of the policy network and critic networks: $\phi$ and $\theta_i, i=1,2$\;
    \For{episode$=1,2,\cdots, E$}{
    Initialize the environment, obtain initial state $s_0$\;
    \For{$t=1,2,\cdots, T$}{
        Select action from the distribution of policy $a_t \sim \pi_{\phi}(a_t|s_t)$\;
        Calculate the user association matrix $\boldsymbol{X(t)}$ by using \textbf{Algorithm \ref{alg:two-phase}}\;
        Execute action $a_t$, obtain reward $r_t$, observe next state $s_{t+1}$\;
        Push transition $(s_t, a_t, r_t, s_{t+1})$ into the replay buffer $\boldsymbol{R}$\;
        Sample a random batch of transitions from $\boldsymbol{R}$\;
        Update critic network parameters by minimize the Eq.~\eqref{eq:lossQ}:\
        $\theta_i \leftarrow \nabla_{\theta_i}J_Q(\theta_i)$, $i \in \{1,2\}$\;
        Update policy network parameters by minimize the Eq.~\eqref{eq:lossp}:\
        $\phi \leftarrow \nabla_{\phi}J_{\pi}(\phi)$\;
        Update target network parameters:\
        $\overline{\theta}_i \leftarrow \varsigma \theta_i + (1-\varsigma)\overline{\theta}_i$, $i \in \{1,2\}$,\
        where $\varsigma$ is the soft update parameter.
    }
    
    }
\end{algorithm}

\par The SAC-based method mainly involves the generation of transitions and the updating of all neural networks, as shown in Algorithm~\ref{alg:sac-tma}. In the initial phase of the training process, the parameters of the policy network and the critic networks are randomly initialized. Following this, an experience replay buffer is constructed. For a certain number of initial steps, the agent obtains the state information $s_t$ and samples a random action $a_t$ from the action space $\mathcal{A}$. After a sufficient number of transitions have been obtained, they are stored in the buffer. Subsequently, the neural networks begin to train, with the policy network outputting actions.

\par During the training period, the agent is required to observe the state information $s_t$ from the environment and to execute the action $a_t$ output by the policy network at each time step $t$, updating the position of UAVs and the transmit power of GUs. Subsequently, the agent receives a instantaneous reward $r_t$ and observes the new environment state $s_{t+1}$, followed by the new transition $(s_t, a_t, r_t, s_{t+1})$ which is stored in the buffer. At each time step, the parameters of neural networks are updated through the sampling of a batch of transitions from the replay buffer. At the same time, the critic networks are updated through the minimization of the loss function in Eq. \eqref{eq:lossQ}, and the policy network is updated through the minimization of the Eq. \eqref{eq:lossp}. Eventually, the target networks are updated.

\par Based on the SAC approach and the formulated MDP, appropriate actions can be output in real-time to optimize the decision variables correspond to the UAV movement and user transmit power. Based on this, and combined with TMA strategy, it is possible to simultaneously optimize user association, UAV movement, and user transmit power to minimize MEC system sum latency and increase the amount of collected data.

\par The main step of SAC-TMA is summarized in Algorithm~\ref{alg:sac-tma}. It can be seen that in SAC-TMA, the user association strategy mainly works by being embedded in the SAC-based method. Specifically, at each time step, the policy network outputs the actions about the movement of UAVs and the transmit power of GUs. Algorithm~\ref{alg:two-phase} is then used to calculate the user association through the updated positions of UAVs and transmit power of GUs, and the instantaneous reward is subsequently calculated. The advantage of this embedding strategy is that it utilizes only part of the environment information of the state in the MDP, which can be regarded as one of the actions output by the policy network, without the need for additional training of the neural network as well as additional information, thus reducing the complexity of training. In addition, since the matching-based algorithm can calculate a stable solution through iteration, the stability of this part of the policies is guaranteed, greatly reducing the training process.

\par For problem $\mathcal{P}$, the SAC-TMA algorithm can be trained by the abovementioned process to be deployed in various computing centers. Specifically, the SAC-TMA algorithm can be iteratively trained for a certain number of episodes until the cumulative rewards stabilizes around a constant value. Afterward, the well-trained algorithm is deployed in a central processing station, such as satellite server, airship, or local server. Moreover, the SAC-TMA algorithm can also be trained online if needed.

\subsection{Complexity Analysis}

\par In this section, the computational complexity of the proposed TMA strategy and the SAC-TMA algorithm is analyzed.
\par \textbf{Complexity of TMA strategy.} In \textit{Phase I}, considering the worst case in Algorithm~\ref{alg:distance based}, each GU needs to send a request to all UAVs, and the complexity is $\mathcal{O}(MN_U)$. Subsequently, the MEC-UAV needs to compare the quality of $N_u^{max}$ GUs, and the complexity is $\mathcal{O}(N_U N_u^{max})$. In the meantime, DC-UAV needs to compare the distance of $N_u^{max}$ GUs, and the complexity is $\mathcal{O}(N_u^{max})$. As such, the total complexity of Algorithm~\ref{alg:distance based} is $\mathcal{O}(N_U \times (M+N_u^{max})+N_u^{max})$, which can be approximated as $\mathcal{O}(M N_U)$. Similarly, the complexity of Algorithm~\ref{alg:rate based} is $\mathcal{O}(M N_U)$. In \textit{Phase II}, considering the worst case, each iteration requires traversing all swap pairs, that is, at most $(N_U-1) {N_u^{max}}^2 N_U +(N-N_u^{max}) N_u^{max}$ swap pairs need to be checked, and the complexity is $\mathcal{O}(N_U^2 + N)$. Given the iteration number $I_L$, the complexity is $\mathcal{O}(I_L N_U^2)$. Eventually, the total complexity of Algorithm~\ref{alg:two-phase} is $\mathcal{O}(I_L N_U^2+M N_U)$.

\par \textbf{Complexity of SAC-TMA}. Define $L_a$ and $L_c$ as the number of hidden layers within the actor and critic networks, respectively. Correspondingly, $A_n$ and $A_c$ indicate the neuron count per layer for the actor and critic networks, respectively, and $B$ denotes the batch size. Therefore, the complexity of the process of updating the actor and the critics is $\mathcal{O}(BL_aA_n^2)$ and $\mathcal{O}(BL_cA_c^2)$, respectively. Thus, the complexity of the training process of the SAC-TMA is $\mathcal{O}(BET(L_a+L_c)(A_n^2+A_c^2))$, where $E$ and $T$ represent the number of episodes and the time steps count per episode, respectively.


\section{Simulation Results and Analysis}
\label{sec:experiments_and_analysis}

\par In this section, we evaluate the performance of the proposed SAC-TMA in addressing the formulated optimization problem.

\subsection{Simulation Configuration}

\par In this section, the system parameter settings and the baseline algorithms are provided.

\begin{table}[tbp]
\caption{Simulation parameters}
\label{tab:parameter}
\setlength{\tabcolsep}{12pt}
\begin{tabular}{@{}ll@{}}
\toprule
Parameters                  & Value                                            \\ \midrule
Maximum number of served users of UAVs $N_u^{max}$           &  \textbf{4}    \\
Task density coefficient $\delta_g$         & 0.2      \\
Length of deadline for each task $D_{m,f}$         & 20 s \\
Noise power spectrum density $n_0$          & $-140$ dBm/Hz                \\
Excessive propagation losses $\eta_{LoS}$, $\eta_{NloS}$ & 0.1, 21            \\
Maximum user transmit power $p_{max}$           & 0.5 W                \\
MEC and DC transmission rate threshold   &1.6 Mbps, 1 Mbps\\
$R_{Mth}$, $R_{Dth}$        &           \\
Bandwidth $W$           & 3 MHz                                            \\
Required CPU cycles per bit data $C$                & 1000 cycles/bit   \\
Effective switching capacitance of MEC-UAVs $\kappa$   & $10^{-27}$     \\

CPU running frequency $\omega$  & $6\times 10^9$ cycles/s         \\
Maximum energy limit for each UAV $E^{max}$    &30 KJ \\ \bottomrule
\end{tabular}%
\end{table}

\begin{table}[tbp]
\caption{Network configurations}
\label{tab:network}
\setlength{\tabcolsep}{27pt}
\begin{tabular}{@{}lll@{}}
\toprule
Parameters                  & Value                                          &  \\ \midrule
Network structure for actor    & [256, 128]                                &  \\
Network structure for critic   & [256, 128]                                        &  \\
Total episodes $E$                 & 5000                                          &  \\
Time step in each episode $T$     & 300                               &  \\
Discount factor $\gamma$       & 0.9                                          &  \\
Target soft update coefficient $\tau$     & 0.005       &  \\
Learning rate for actor        & $3\times 10^{-4}$                                         &  \\
Learning rate for critic       & $10^{-4}$                                   &  \\
Replay buffer size             & $10^6$                                        &  \\
Entropy regularization coefficient     & 0.2                                           &  \\
Batch size                      & 256                                            &  \\
\bottomrule
\end{tabular}%
\end{table}

\begin{figure*}[!t]
\centering
\includegraphics[width= \textwidth]{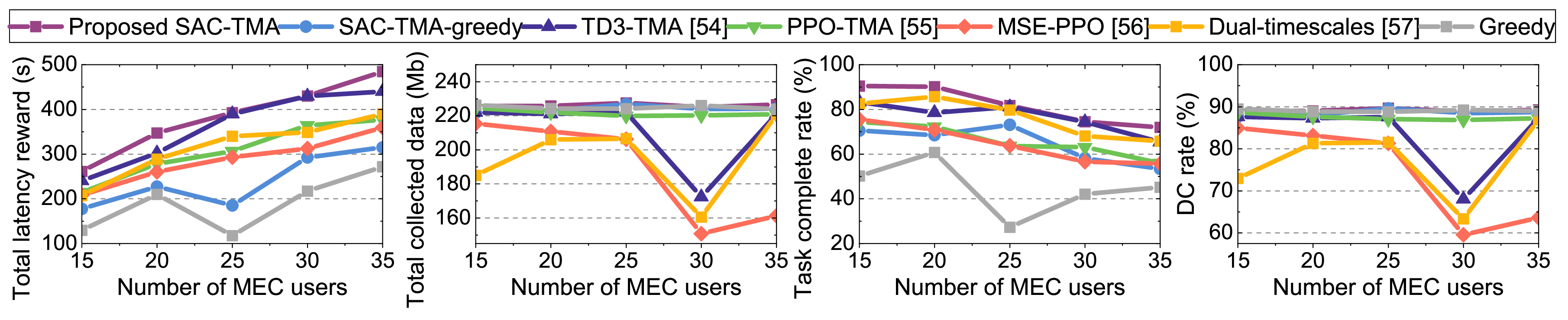}
\caption{Effect of MEC users number on latency and DC performance (The performance of latency is measured by the latency reward defined in Eq.~\eqref{eq:latency reward}).}
\label{fig:mec user}
\end{figure*}

\begin{figure}[!t]
\centering
\includegraphics[width= 0.49\textwidth]{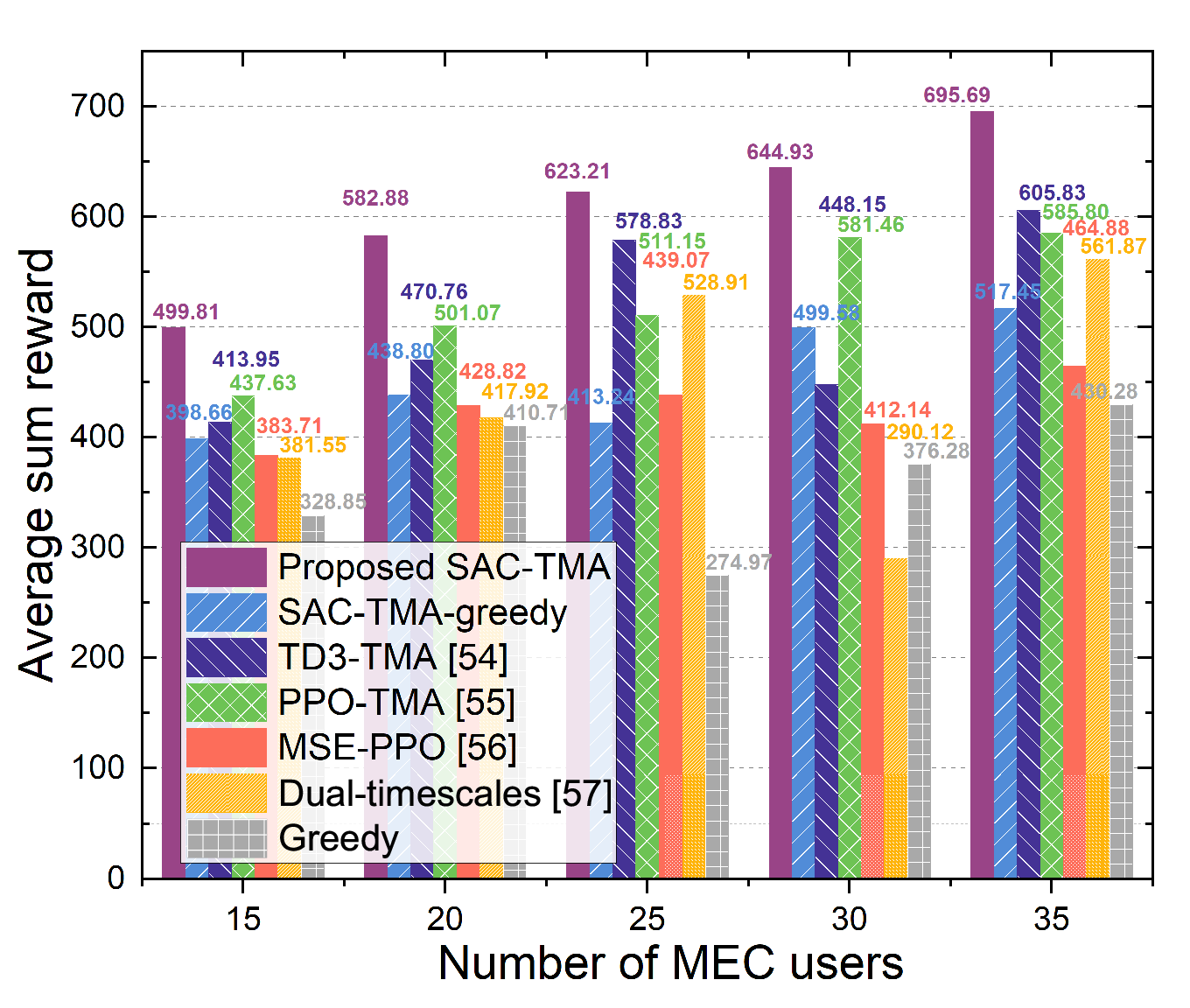}
\caption{Sum reward under different numbers of MEC users with $N=10$.}
\label{fig:sumreward}
\end{figure}

\par \textit{1) System Parameter Settings:} In the simulation, a square area is considered, with the size of $1500\times 1500$ m\textsuperscript{2}, and $(x_{min}, y_{min})=(-750,-750)$, $(x_{max},y_{max})=(750,750)$. In addition, due to the significant influence of interference in the joint network, we set the initial positions of UAVs as (-500, 500), (-500, -500), (500, 500), and (500, -500) to avoid strong initial interference. The number of UAVs is $M=4$, with 3 MEC-UAVs and 1 DC-UAV. Moreover, the number of GUs is set to 35 including 25 MEC-users and 10 DC-users, and the GUs are stationary and randomly distributed over the region under consideration. In addition, the altitude of UAVs is maintained at a fixed level of $H = 100$ m, the safety distance between UAVs is 3 m and the maximum flight speed is 50 m/s \cite{Zhan2021}. For the MEC subsystem, there are three types of tasks, each GU randomly selects one type to generate from the task set $\{F_1, F_2, F_3\}$, where $F_1=512$ Kbits, $F_2=256$ Kbits, $F_3=128$ Kbits, and the corresponding probability set is $\{P_1, P_2, P_3\}$, where $P_1=0.2$, $P_2=0.3$, $P_3=0.5$. Accordingly, the latency limit set of the tasks is $\{T_1, T_2, T_3\}$, where $T_1= 1$ s, $T_2= 0.5$ s, $T_3=0.25$ s. For the DC subsystem, the storage limit of each DC-user $n$ is $L_n^{max}=60$ Mbits, and for simplicity, the data generation of the DC subsystem uses the same set as the MEC task set. Besides, the remaining system parameters are presented in Table~\ref{tab:parameter}, in which the parameters are configured according to \cite{Chen2023}, \cite{Yu2021}, and \cite{Mu2021}. Moreover, the network structure and parameters of SAC-TMA and other benchmark algorithms are given in Table~\ref{tab:network}. Furthermore, to reflect the performance after convergence, all simulation results are averages of data after 1000 episodes.

\begin{figure*}[!t]
    \centering
    \includegraphics[width=\textwidth]{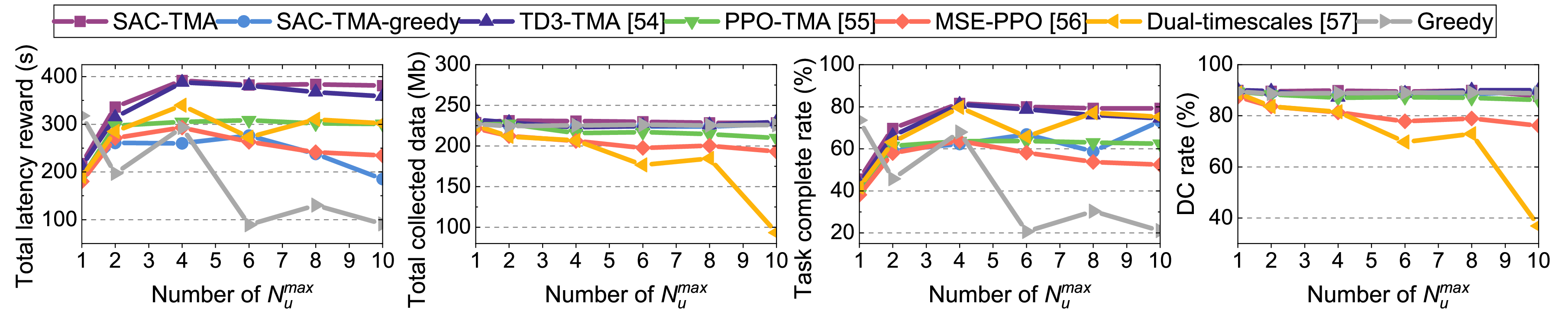}
    \caption{Effect of $N_u^{max}$ on latency reward, DC reward, task complete rate, and DC rate.}
    \label{fig:overall_performance}
\end{figure*}

\begin{figure}[!t]
\centering
\includegraphics[width=0.49\textwidth]{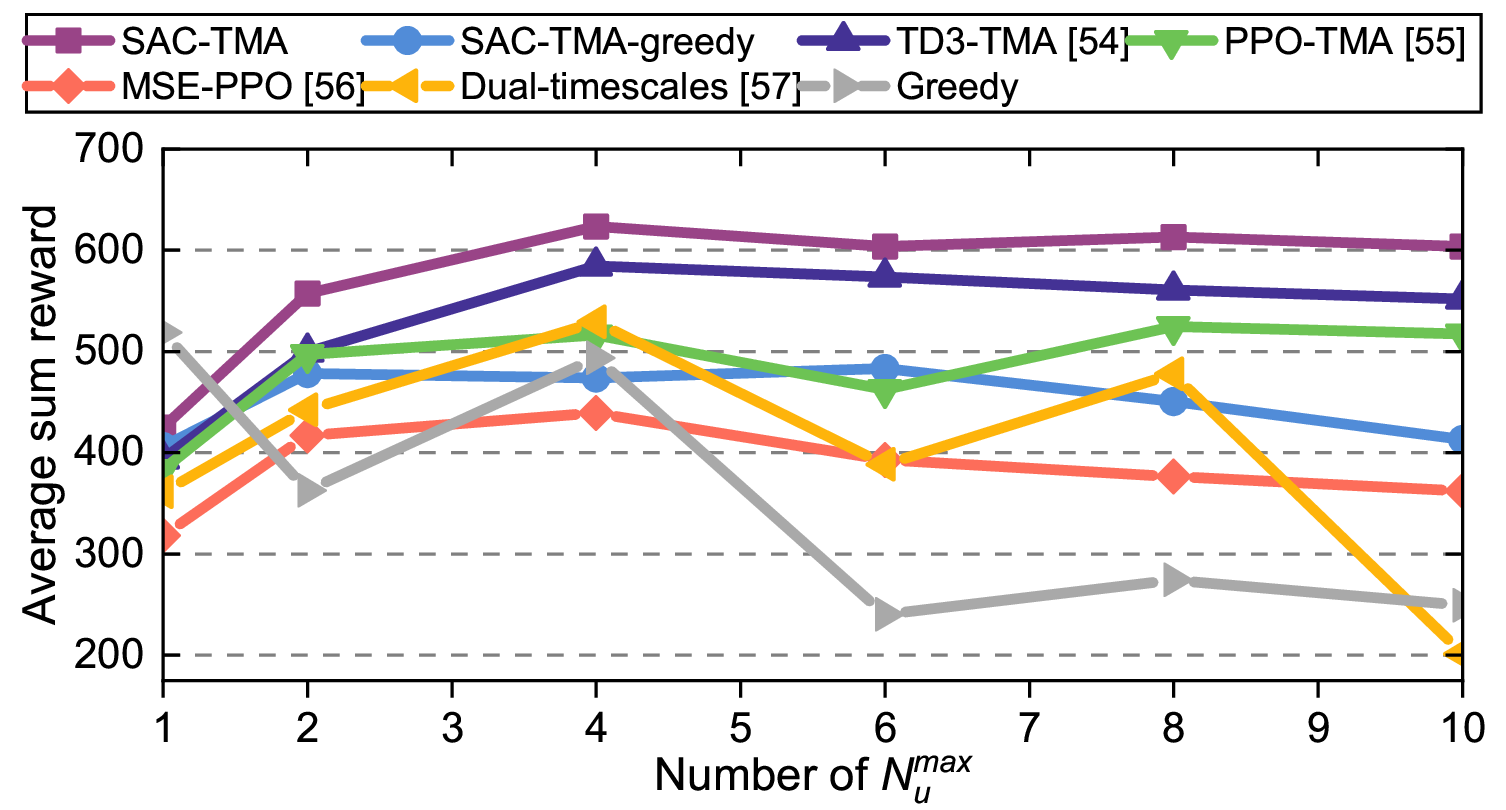}
\caption{Sum reward under different numbers of maximum service capacity of one UAV $N_u^{max}$.}
\label{fig:max user sum reward}
\end{figure}

\par \textit{2) Baseline Algorithms:} In this work, the proposed DRL-based approach is compared with four other benchmark algorithms which are SAC-TMA-greedy, twin delayed deep deterministic policy gradient (TD3)-based algorithm~\cite{td32018}, proximal policy optimization (PPO)-based algorithm~\cite{ppo2017}, maximize service efficiency proximal policy optimization (MSE-PPO) algorithm~\cite{li2024uav}, dual-timescales optimization scheme~\cite{Huang2024}, and distance-based greedy algorithm, and the details of benchmark algorithms are as follows:

\begin{itemize}
    \item \textit{Distance-Greedy Algorithm:}~The greedy approach preferentially selects the closest GU to perform task offloading as well as DC and uses Algorithm~\ref{alg:distance based} for user association. In addition, the power of each GU is assigned by random generation.
    \item \textit{SAC-TMA-greedy:}~A variant of the SAC algorithm and greedy algorithm, where the SAC algorithm integrates the proposed TMA strategy to control the power allocation and trajectory planning for MEC-UAVs and user association for all UAVs to perform task offloading and DC, and the greedy algorithm only controls the trajectory of the DC-UAV.
    \item \textit{TD3-TMA} \cite{td32018}: A variant of the TD3 algorithm integrates the proposed TMA strategy and centrally controls the UAV trajectory, power allocation, and user association for task offloading of MEC-UAVs and DC of the DC-UAV.
    \item \textit{PPO-TMA} \cite{ppo2017}: A variant of the PPO algorithm integrates the proposed TMA strategy and centrally controls the UAV trajectory, power allocation, and user association for task offloading of MEC-UAVs and DC of the DC-UAV.
    \item \textit{MSE-PPO} \cite{li2024uav}: This paper proposes a parametrized, parallel actor structure-based method named maximize service efficiency proximal policy optimization (MSE-PPO) algorithm to update task offloading and flight hybrid policy parameters separately. In the simulations, the sub-actor with discrete action space controls the user association and sub-actors with continuous action spaces control the UAV trajectory and power allocation.
    \item \textit{Dual-timescales} \cite{Huang2024}: This paper proposes a dual-timescales optimization scheme for joint resource slicing and task scheduling. On small timescales, a self-attention mechanism-based TD3 algorithm improves the negative impact of extreme actions. On large timescales, a heuristic-based artificial electric field (AEF) approach obtains a resource slicing policy. Since the iterative process of the AEF approach is quite time consuming in our considered optimization problem, we replace the AEF method with Algorithm~\ref{alg:distance based} in large timescales to control the user association and use self-attention mechanism-based TD3 algorithm in small timescales to control the UAV trajectory and power allocation.
\end{itemize}

\subsection{Simulation Results}
\par In this section, the performance of the proposed SAC-TMA as well as the benchmark algorithms are evaluated and analyzed in terms of comparison results, convergence results, and trajectory results, respectively. Moreover, we perform an effectiveness analysis of the proposed TMA strategy.

\subsubsection{Comparison Results}

\par In this part, the performance of the SAC-TMA algorithm and baseline algorithms is evaluated in terms of system latency, collected data, MEC task completion rate, DC rate, and average cumulative rewards under different numbers of MEC-users and $N_u^{max}$. Moreover, the average energy consumption of the UAVs when executing different algorithms is given.

\par Figs. \ref{fig:mec user} and \ref{fig:sumreward} illustrate the MEC, DC, and overall performance obtained by our proposed SAC-TMA algorithm and other benchmark approaches with different numbers of MEC users. It can be observed from Fig. \ref{fig:mec user} that as the number of MEC users increases, the total latency rewards rise while the task complete rates fall. This may be due to more MEC tasks allowing for higher rewards but also causing more severe interference and UAV scheduling problems, resulting in a decrease in the task complete rate. Moreover, when the number of MEC users reaches 30, the DC performance of TD3-TMA, MSE-PPO, and Dual-timescales methods decreases significantly, while the SAC-TMA algorithm outperforms in terms of all key metrics of MEC and DC and sum rewards. Additionally, it can be observed from Fig. \ref{fig:sumreward} that our proposed SAC-TMA algorithm exhibits high average cumulative rewards for all MEC user numbers, which reflects its better stability and adaptability on the considered optimization problem.

\par Figs. \ref{fig:overall_performance} and \ref{fig:max user sum reward} illustrate the total latency reward, the volume of collected data, the MEC task completion rate, the DC rate, and the sum rewards under different numbers of $N_u^{max}$. It can be observed that the proposed SAC-TMA algorithm outperforms other learning-based and greedy algorithms in all cases, demonstrating the stability of SAC-TMA. Moreover, all learning-based baseline algorithms integrated with TMA present superior performance in terms of sum rewards and DC performance compared to other algorithms. This may be because the TMA strategy can output stable matching policies to mitigate interference, allowing the agent to achieve high MEC rewards while maintaining a better DC performance. Particularly, we observe that when $N_u^{max}=4$, all algorithms can reach their peak performance in various aspects. Therefore, we set $N_u^{max}$ to 4 in the simulation parameter settings.

\par
\begin{figure}[!t]
\centering
\includegraphics[width=0.49\textwidth]{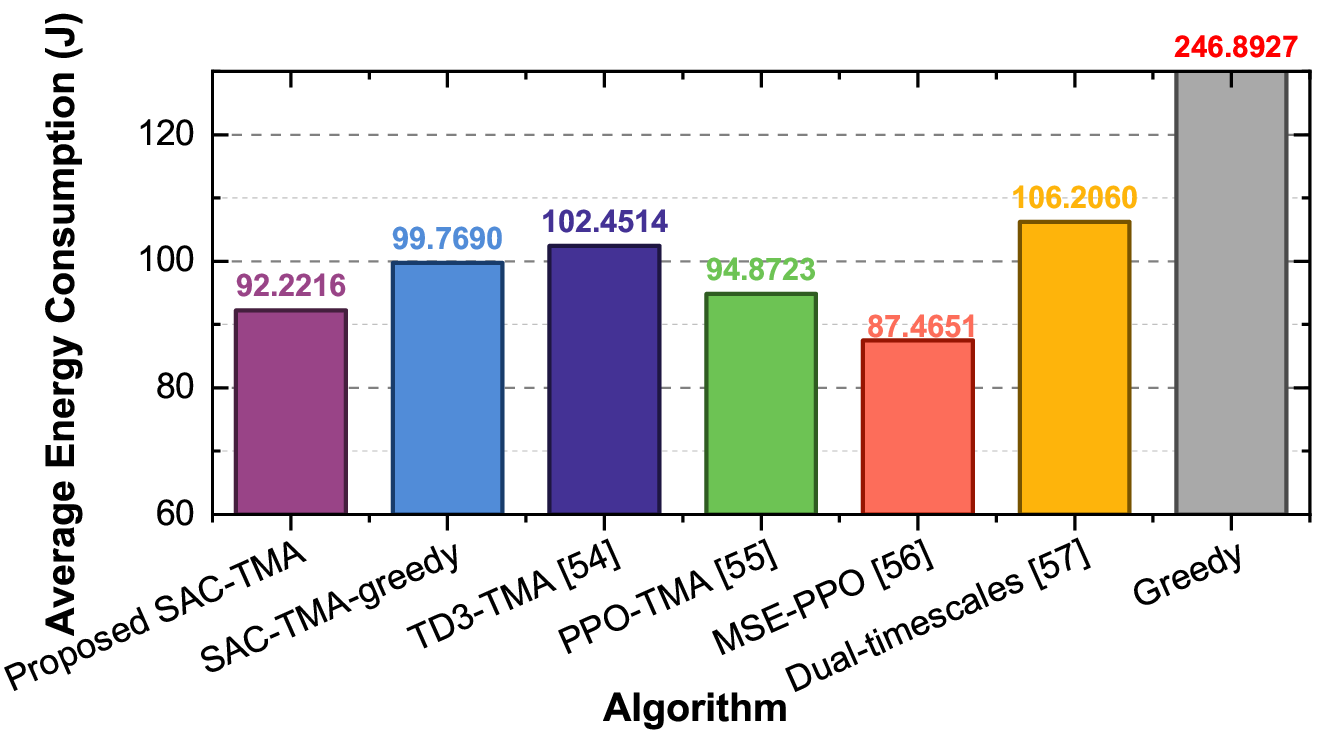}
\caption{Average energy consumption of one UAV during a step.}
\label{fig:energy}
\end{figure}

 In addition, Fig.~\ref{fig:energy} presents the average energy consumed by each UAV at each time step under different algorithms. As can be seen, the average energy consumption of SAC-TMA is lower than that of the other five algorithms except for MSE-PPO, which are 92.44\%, 90.01\%, 97.21\%, 86.83\%, and 37.35\% of the SAC-TMA-greedy, TD3-TMA, PPO-TMA, Dual-timescales, and greedy algorithms, respectively. This may be because MSE-PPO learns policies that favor less movement or fewer associations during the training process, thereby consuming less energy, while resulting in fewer sum rewards. Moreover, the average energy consumption using learning-based algorithms is significantly lower than that of the traditional algorithm using greedy, which indicates that thanks to the energy penalty in Eq. \eqref{eq:energy penalty}, UAVs will consider energy saving while ensuring system performance.

\begin{figure}[!t]
\centering
\includegraphics[width= 0.45\textwidth]{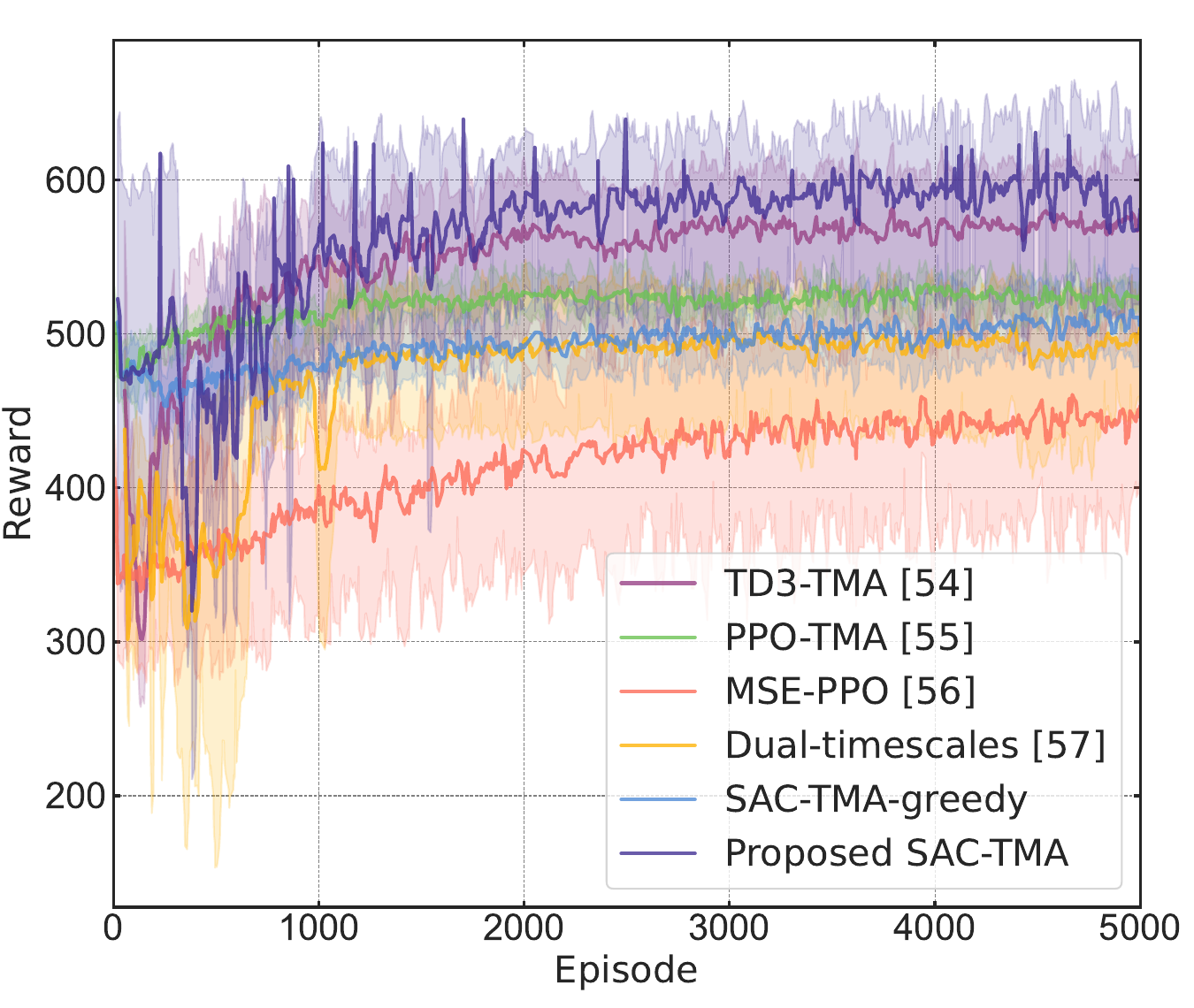}
\caption{Training reward using random seeds 0, 1, and 2 (The curves represent the mean level and the shaded areas represent the range of standard deviation).}
\label{fig:reward}
\end{figure}

\begin{figure}[!t]
\centering
\includegraphics[width= 0.5\textwidth]{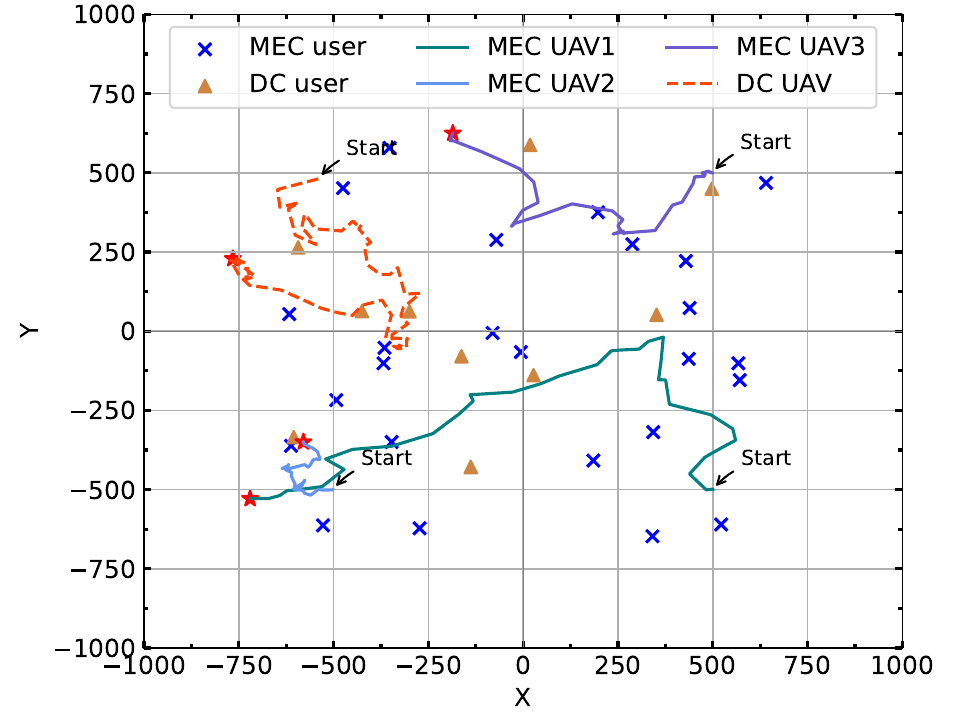}
\caption{UAV movement trajectory with $M=25$, $N=10$ (The start and end points are marked with text and red stars, respectively).}
\label{fig:trajectory}
\end{figure}

\subsubsection{Convergence Results}

\par In this part, the convergence performance of the learning-based algorithms is investigated.

\par Fig.~\ref{fig:reward} illustrates the cumulative reward curves of different learning-based algorithms. As illustrated in Fig.~\ref{fig:reward}, after approximately 1000 episodes of training, the SAC-TMA algorithm tends to converge, and it exhibits a gradual upward trend in the later stages, with an overall reward superior to other algorithms, demonstrating the advantage of long-term performance. This may be due to the fact that the SAC-TMA algorithm adopts a strategy that maximizes the cumulative reward and policy entropy, which enables the agent to enhance exploration and continuously learn new policies. At the same time, the structure of double critic can prevent the problem of excessive Q-value. Specifically, to reduce the bias caused by overestimation, SAC-TMA does not directly use the estimated value of a single Q-network but selects the minimum value derived from the estimations of two Q-networks for calculating the target Q-value. This effectively avoids the strategy failure caused by high estimates from a single Q-network, thereby enhancing the stability of SAC-TMA. Moreover, it can be observed that learning-based algorithms with the TMA strategy exhibit better performance compared with other methods, even though they may originate from the same baseline algorithm. This may be primarily attributed to the effectiveness of the proposed TMA strategy in handling user association, which can allow the agent to achieve high MEC rewards while maintaining better DC performance.

\subsubsection{Trajectory Results}

\par This part presents an evaluation of the effectiveness of the SAC-TMA algorithm based on an analysis of the movement trajectories of the UAVs. Fig. \ref{fig:trajectory} shows the movement trajectories of 4 UAVs within 300 time steps. It can be observed that the trajectories of UAVs are able to cover most of the GUs, and all the UAVs tend to move toward the locations where GUs gather, and there is no collision. This may be due to the algorithm learning strategies that keep UAVs at a distance to mitigate mutual interference. Moreover, the DC-UAV tends to move away from the MEC-UAVs. This may be because the agent adopts a strategy that sacrifices smaller DC rewards to avoid severe interference with MEC users, thereby balancing the income of rewards.

\subsubsection{Effectiveness Analysis}

\par In this part, we compare the proposed TMA strategy with five other strategies base on the utility in Eq.~\eqref{eq:utility}, which is related to the system sum rate. Moreover, the running times of each strategy are compared to analyze the feasibility in terms of computational complexity, and the details are as follows:

\begin{itemize}
    \item Random generated matching strategy (\textit{Random}): The strategy for randomly generating matching.
    \item Distance-based stable matching strategy (\textit{Distance-based}): The GS-based strategy using distance-based evaluation, which is shown in Algorithm 1.
    \item Distance-rate-based two-step stable matching strategy (\textit{Distance-rate-based}): The GS-based strategy using rate-based evaluation, which is shown in Algorithm 2.
    \item Random initialization swap strategy (\textit{Swap}): The conventional swap strategy which is initialized with randomly generated matching.
    \item Distance-based initialization swap strategy (\textit{Distance-swap}): The enhanced swap strategy initialized by using distance-based pre-evaluation.
\end{itemize}

\par At each time step, the movement direction and distance for all UAVs are randomly generated, focusing solely on user association to calculate the system sum rate and the execution time of the association algorithms. Moreover, to reduce the deviation of randomness, we use three different random seeds, 0, 1, and 2, and take the average values under the three random seeds as the final results. 

\par
\begin{figure}[!t]
\centering
\includegraphics[width= 0.5\textwidth]{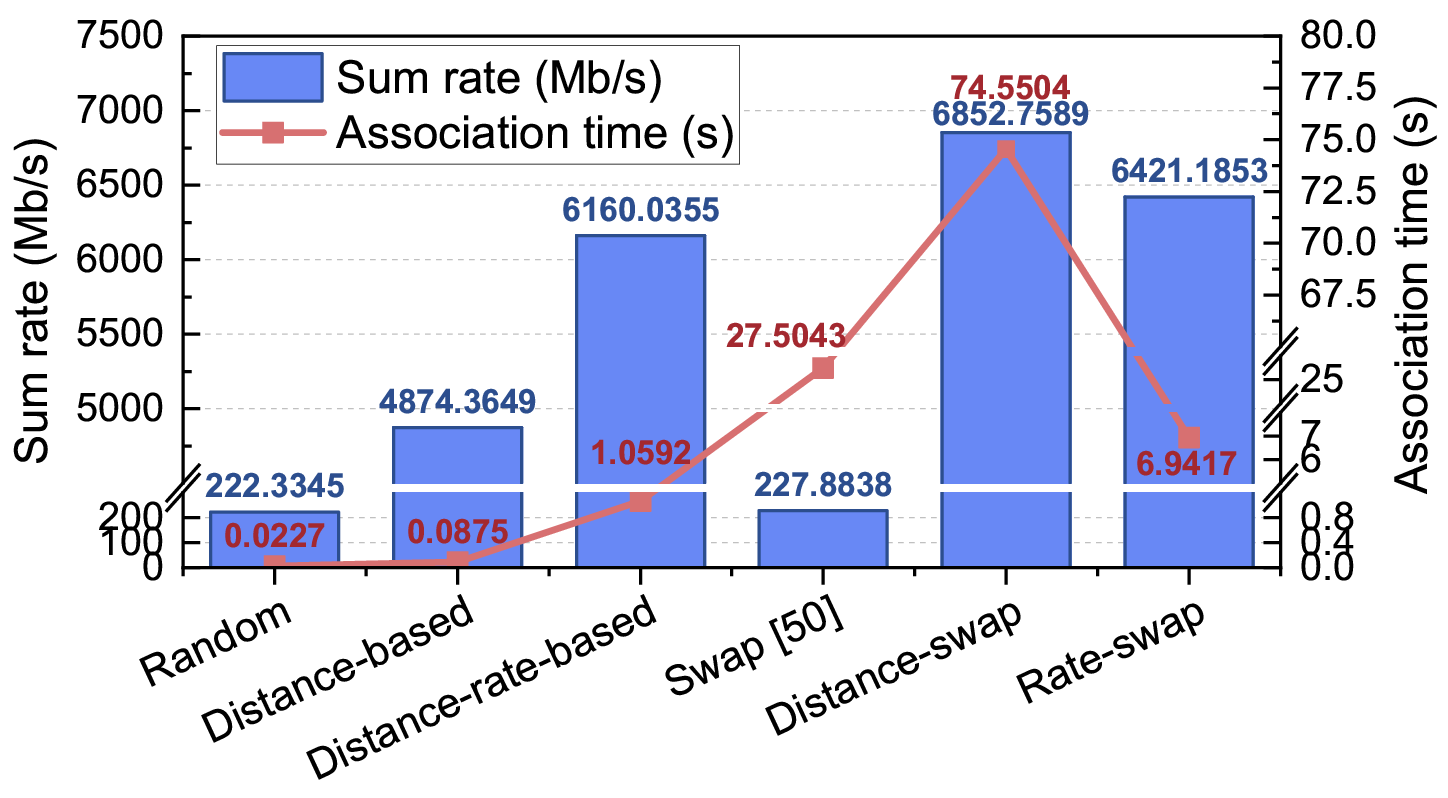}
\caption{Comparison of system sum rate and total association time for 300 runs of different association algorithms (Breakpoints processed).}
\label{fig:association}
\end{figure}

 Fig.~\ref{fig:association} illustrates the comparison results of the system sum rate and running times of different strategies. As can be seen, the performance of conventional algorithms using random and GS concepts is inferior to their swap version. This is due to the incremental principle of the swap operation, which means that the swap operation is only performed when the performance of the solution will improve. Therefore, the solution obtained by the swap algorithm will not be inferior to any of the previous solutions. Moreover, the swap algorithm using the pre-evaluation method significantly outperforms the swap method with random initialization. This is because the pre-evaluation scheme utilizes environmental information, making the initial solution superior to the random solution.

\par It is found that the original version of the swap algorithm has the relative worst overall performance (only 2.5\% improvement over random association, while the running time increased by 1200 more times), which may be due to the poor initial quality and large solution search space. Among the two-phase swap algorithms, the distance-swap algorithm has the best association effect. Nevertheless, the running time of the distance-based algorithm is intolerable. Note that the performance of the rate-swap algorithm is the best except for the distance-swap algorithm, and the average running time of each slot is about $2\times10^{-2}$ s, which is significantly shorter than the duration of a slot, such as 1 s. Hence, it is practical in terms of the running time. Therefore, we choose the rate-swap algorithm, namely the TMA strategy, as the association algorithm in this work.

\section{Conclusion}\label{sec:conclusion}

\par This paper has investigated a multi-UAV-assisted joint MEC-DC uplink communication system. Specifically, we have considered a joint MEC-DC scenario consisting of a multi-UAV-assisted MEC subsystem and a single-UAV-assisted DC subsystem, and formulated a joint optimization problem to minimize the MEC system sum latency and maximize the volume of collected data. Based on this, the problem has been reformulated as an MDP, and a one-to-many matching game has been modeled to simplify the optimization of decision variables, thus improving the training efficiency of the DRL algorithm. Then, we have proposed the SAC-TMA to obtain real-time feasible policies. Simulation results have demonstrated that the proposed SAC-TMA algorithm is effective in reducing the MEC latency while improving the volume of collected data compared with other benchmark algorithms.
\bibliographystyle{IEEEtran}
\bibliography{bib}

\begin{thebibliography}{10}
\providecommand{\url}[1]{#1}
\csname url@samestyle\endcsname
\providecommand{\newblock}{\relax}
\providecommand{\bibinfo}[2]{#2}
\providecommand{\BIBentrySTDinterwordspacing}{\spaceskip=0pt\relax}
\providecommand{\BIBentryALTinterwordstretchfactor}{4}
\providecommand{\BIBentryALTinterwordspacing}{\spaceskip=\fontdimen2\font plus
\BIBentryALTinterwordstretchfactor\fontdimen3\font minus \fontdimen4\font\relax}
\providecommand{\BIBforeignlanguage}[2]{{%
\expandafter\ifx\csname l@#1\endcsname\relax
\typeout{** WARNING: IEEEtran.bst: No hyphenation pattern has been}%
\typeout{** loaded for the language `#1'. Using the pattern for}%
\typeout{** the default language instead.}%
\else
\language=\csname l@#1\endcsname
\fi
#2}}
\providecommand{\BIBdecl}{\relax}
\BIBdecl

\bibitem{Adnan2024}
M.~H. Adnan, Z.~A. Zukarnain, and O.~A. Amodu, ``Fundamental design aspects of {UAV}-enabled {MEC} systems: {A} review on models, challenges, and future opportunities,'' \emph{Comput. Sci. Rev.}, vol.~51, p. 100615, 2024.

\bibitem{ZheSurvey}
Z.~{Wang}, J.~{Zhang}, H.~{Du}, D.~{Niyato}, S.~{Cui}, B.~{Ai}, M.~{Debbah}, K.~B. {Letaief}, and H.~V. {Poor}, ``A tutorial on extremely large-scale {MIMO} for {6G}: Fundamentals, signal processing, and applications,'' \emph{IEEE Commun. Surveys Tuts.}, vol.~26, no.~3, pp. 1560--1605, 3rd quarter, 2024.

\bibitem{Zhu2024}
Y.~Zhu, B.~Yang, M.~Liu, and Z.~Li, ``{UAV} trajectory optimization for large-scale and low-power data collection: {A}n attention-reinforced learning scheme,'' \emph{{IEEE} Trans. Wirel. Commun.}, vol.~23, no.~4, pp. 3009--3024, 2024.

\bibitem{Du2024}
P.~Du, Y.~Shi, H.~Cao, S.~Garg, G.~Kaddoum, and M.~Alrashoud, ``3-d trajectory optimization and communication resources allocation in {UAV}-assisted {IoT} networks for sustainable industry 5.0,'' \emph{{IEEE} Trans. Consumer Electron.}, vol.~70, no.~1, pp. 1423--1433, 2024.

\bibitem{huang2024a}
J.~Huang, A.~Wang, G.~Sun, J.~Li, J.~Wang, H.~Du, and D.~Niyato, ``Dual {UAV} cluster-assisted maritime physical layer secure communications via collaborative beamforming,'' \emph{IEEE Internet Things J.}, pp. 1--1, 2024.

\bibitem{zhang2024}
C.~Zhang, G.~Sun, Q.~Wu, J.~Li, S.~Liang, D.~Niyato, and V.~C.~M. Leung, ``{UAV} swarm-enabled collaborative secure relay communications with time-domain colluding eavesdropper,'' \emph{{IEEE} Trans. Mob. Comput.}, vol.~23, no.~9, pp. 8601--8619, 2024.

\bibitem{zhang2024a}
C.~Zhang, G.~Sun, J.~Li, Q.~Wu, J.~Wang, D.~Niyato, and Y.~Liu, ``Multi-objective aerial collaborative secure communication optimization via generative diffusion model-enabled deep reinforcement learning,'' \emph{IEEE Trans. Mob. Comput.}, pp. 1--18, 2024.

\bibitem{Wu2021}
Q.~Wu, J.~Xu, Y.~Zeng, D.~W.~K. Ng, N.~Al{-}Dhahir, R.~Schober, and A.~L. Swindlehurst, ``A comprehensive overview on {5G}-and-beyond networks with {UAV}s: {F}rom communications to sensing and intelligence,'' \emph{{IEEE} J. Sel. Areas Commun.}, vol.~39, no.~10, pp. 2912--2945, 2021.

\bibitem{Jiang2023}
X.~Jiang, M.~Sheng, N.~Zhao, J.~Liu, D.~Niyato, and F.~R. Yu, ``Outage analysis of {UAV}-aided networks with underlaid ambient backscatter communications,'' \emph{{IEEE} Trans. Wirel. Commun.}, vol.~22, no.~11, pp. 7492--7505, 2023.

\bibitem{pei2022}
J.~Pei, H.~Chen, and L.~Shu, ``{UAV}-assisted connectivity enhancement algorithms for multiple isolated sensor networks in agricultural {I}nternet of things,'' \emph{Comput. Networks}, vol. 207, p. 108854, 2022.

\bibitem{Zhou2022a}
M.~Zhou, H.~Chen, L.~Shu, and Y.~Liu, ``{UAV}-assisted sleep scheduling algorithm for energy-efficient data collection in agricultural {I}nternet of things,'' \emph{{IEEE} Internet Things J.}, vol.~9, no.~13, pp. 11\,043--11\,056, 2022.

\bibitem{Masuduzzaman2022}
M.~Masuduzzaman, A.~Islam, K.~Sadia, and S.~Y. Shin, ``{UAV}-based {MEC}-assisted automated traffic management scheme using blockchain,'' \emph{Future Gener. Comput. Syst.}, vol. 134, pp. 256--270, 2022.

\bibitem{Elloumi2018}
M.~Elloumi, R.~Dhaou, B.~Escrig, H.~Idoudi, and L.~A. Sa{\"{\i}}dane, ``Monitoring road traffic with a {UAV}-based system,'' in \emph{Proc. {IEEE} {WCNC}}, 2018, pp. 1--6.

\bibitem{sun2024}
G.~Sun, L.~He, Z.~Sun, Q.~Wu, S.~Liang, J.~Li, D.~Niyato, and V.~C.~M. Leung, ``Joint task offloading and resource allocation in aerial-terrestrial {UAV} networks with edge and fog computing for post-disaster rescue,'' \emph{{IEEE} Trans. Mob. Comput.}, vol.~23, no.~9, pp. 8582--8600, 2024.

\bibitem{dong2021}
J.~Dong, K.~Ota, and M.~Dong, ``Uav-based real-time survivor detection system in post-disaster search and rescue operations,'' \emph{IEEE Journal on Miniaturization for Air and Space Systems}, vol.~2, no.~4, pp. 209--219, 2021.

\bibitem{zheng2024}
X.~Zheng, G.~Sun, J.~Li, S.~Liang, Q.~Wu, M.~Yin, D.~Niyato, and V.~C.~M. Leung, ``Reliable and energy-efficient communications via collaborative beamforming for {UAV} networks,'' \emph{{IEEE} Trans. Wirel. Commun.}, vol.~23, no.~10, pp. 13\,235--13\,251, 2024.

\bibitem{sun2024a}
G.~Sun, Y.~Wang, Z.~Sun, Q.~Wu, J.~Kang, D.~Niyato, and V.~C.~M. Leung, ``Multi-objective optimization for multi-uav-assisted mobile edge computing,'' \emph{{IEEE} Trans. Mob. Comput.}, vol.~23, no.~12, pp. 14\,803--14\,820, 2024.

\bibitem{sun2024b}
Z.~Sun, G.~Sun, Q.~Wu, L.~He, S.~Liang, H.~Pan, D.~Niyato, C.~Yuen, and V.~C.~M. Leung, ``{TJCCT}: {A} two-timescale approach for {UAV}-assisted mobile edge computing,'' \emph{IEEE Trans. Mob. Comput.}, pp. 1--18, 2024.

\bibitem{Li2024a}
B.~Li, R.~Yang, L.~Liu, J.~Wang, N.~Zhang, and M.~Dong, ``Robust computation offloading and trajectory optimization for multi-{UAV}-assisted {MEC:} {A} multiagent {DRL} approach,'' \emph{{IEEE} Internet Things J.}, vol.~11, no.~3, pp. 4775--4786, 2024.

\bibitem{liu2024b}
L.~Liu, A.~Wang, G.~Sun, J.~Li, H.~Pan, and T.~Q.~S. Quek, ``Multi-objective optimization for data collection in {UAV}-assisted agricultural {IoT},'' \emph{IEEE Trans. Veh. Technol.}, pp. 1--17, 2024.

\bibitem{Sun2023}
C.~Sun, X.~Xiong, Z.~Zhai, W.~Ni, T.~Ohtsuki, and X.~Wang, ``Max-min fair {3D} trajectory design and transmission scheduling for solar-powered fixed-wing {UAV}-assisted data collection,'' \emph{{IEEE} Trans. Wirel. Commun.}, vol.~22, no.~12, pp. 8650--8665, 2023.

\bibitem{sun2023a}
J.~Sun, G.~Xu, T.~Zhang, X.~Yang, M.~Alazab, and R.~H. Deng, ``Privacy-aware and security-enhanced efficient matchmaking encryption,'' \emph{{IEEE} Trans. Inf. Forensics Secur.}, vol.~18, pp. 4345--4360, 2023.

\bibitem{sun2024privacy}
J.~Sun, Y.~Bao, W.~Qiu, R.~Lu, S.~Zhang, Y.~Guan, and X.~Cheng, ``Privacy-preserving fine-grained data sharing with dynamic service for the cloud-edge {IoT},'' \emph{IEEE Trans. Dependable Secure Comput.}, 2024.

\bibitem{Li2024}
H.~Li, J.~Zhang, H.~Zhao, Y.~Ni, J.~Xiong, and J.~Wei, ``Joint optimization on trajectory, computation and communication resources in information freshness sensitive {MEC} system,'' \emph{{IEEE} Trans. Veh. Technol.}, vol.~73, no.~3, pp. 4162--4177, 2024.

\bibitem{Chen2023}
J.~Chen, X.~Cao, P.~Yang, M.~Xiao, S.~Ren, Z.~Zhao, and D.~O. Wu, ``Deep reinforcement learning based resource allocation in multi-{UAV}-aided {MEC} networks,'' \emph{{IEEE} Trans. Commun.}, vol.~71, no.~1, pp. 296--309, 2023.

\bibitem{Dandapat2024}
J.~Dandapat, N.~Gupta, S.~Agarwal, and B.~Kumbhani, ``Service time maximization for data collection in multi-{UAV}-aided networks,'' \emph{{IEEE} Trans. Intell. Veh.}, vol.~9, no.~1, pp. 328--337, 2024.

\bibitem{Zhao2024}
T.~Zhao, F.~Li, and L.~He, ``Secure video offloading in multi-uav-enabled {MEC} networks: {A} deep reinforcement learning approach,'' \emph{{IEEE} Internet Things J.}, vol.~11, no.~2, pp. 2950--2963, 2024.

\bibitem{Yang2020}
L.~Yang, H.~Yao, J.~Wang, C.~Jiang, A.~Benslimane, and Y.~Liu, ``Multi-{UAV}-enabled load-balance mobile-edge computing for {IoT} networks,'' \emph{{IEEE} Internet Things J.}, vol.~7, no.~8, pp. 6898--6908, 2020.

\bibitem{Yu2020}
Z.~Yu, Y.~Gong, S.~Gong, and Y.~Guo, ``Joint task offloading and resource allocation in uav-enabled mobile edge computing,'' \emph{{IEEE} Internet Things J.}, vol.~7, no.~4, pp. 3147--3159, 2020.

\bibitem{Zhan2021}
C.~Zhan, H.~Hu, Z.~Liu, Z.~Wang, and S.~Mao, ``Multi-uav-enabled mobile-edge computing for time-constrained {IoT} applications,'' \emph{{IEEE} Internet Things J.}, vol.~8, no.~20, pp. 15\,553--15\,567, 2021.

\bibitem{Zhou2022}
X.~Zhou, L.~Huang, T.~Ye, and W.~Sun, ``Computation bits maximization in {UAV}-assisted {MEC} networks with fairness constraint,'' \emph{{IEEE} Internet Things J.}, vol.~9, no.~21, pp. 20\,997--21\,009, 2022.

\bibitem{Lee2024}
W.~Lee and T.~Kim, ``Multiagent reinforcement learning in controlling offloading ratio and trajectory for multi-uav mobile-edge computing,'' \emph{{IEEE} Internet Things J.}, vol.~11, no.~2, pp. 3417--3429, 2024.

\bibitem{Pervez2024}
F.~Pervez, A.~Sultana, C.~Yang, and L.~Zhao, ``Energy and latency efficient joint communication and computation optimization in a multi-{UAV}-assisted {MEC} network,'' \emph{{IEEE} Trans. Wirel. Commun.}, vol.~23, no.~3, pp. 1728--1741, 2024.

\bibitem{wang2023a}
H.~Wang, H.~Zhang, X.~Liu, K.~Long, and A.~Nallanathan, ``Joint {UAV} placement optimization, resource allocation, and computation offloading for thz band: {A} {DRL} approach,'' \emph{{IEEE} Trans. Wirel. Commun.}, vol.~22, no.~7, pp. 4890--4900, 2023.

\bibitem{Liu2022}
Y.~Liu, J.~Yan, and X.~Zhao, ``Deep reinforcement learning based latency minimization for mobile edge computing with virtualization in maritime {UAV} communication network,'' \emph{{IEEE} Trans. Veh. Technol.}, vol.~71, no.~4, pp. 4225--4236, 2022.

\bibitem{Yu2021}
Y.~Yu, J.~Tang, J.~Huang, X.~Zhang, D.~K.~C. So, and K.~Wong, ``Multi-objective optimization for {UAV}-assisted wireless powered {IoT} networks based on extended {DDPG} algorithm,'' \emph{{IEEE} Trans. Commun.}, vol.~69, no.~9, pp. 6361--6374, 2021.

\bibitem{Liu2024}
K.~Liu and J.~Zheng, ``{UAV} trajectory planning with interference awareness in {UAV}-enabled time-constrained data collection systems,'' \emph{{IEEE} Trans. Veh. Technol.}, vol.~73, no.~2, pp. 2799--2815, 2024.

\bibitem{Zeng2023}
Y.~Zeng and J.~Tang, ``Mec-assisted real-time data acquisition and processing for {UAV} with general missions,'' \emph{{IEEE} Trans. Veh. Technol.}, vol.~72, no.~1, pp. 1058--1072, 2023.

\bibitem{Liu2024a}
J.~Liu, X.~Zhao, P.~Qin, S.~Geng, Z.~Chen, and H.~Zhou, ``Learning-based multi-{UAV} assisted data acquisition and computation for information freshness in {WPT} enabled space-air-ground {PIoT},'' \emph{{IEEE} Trans. Netw. Sci. Eng.}, vol.~11, no.~1, pp. 48--63, 2024.

\bibitem{seid2021}
A.~M. Seid, G.~O. Boateng, B.~Mareri, G.~Sun, and W.~Jiang, ``Multi-agent {DRL} for task offloading and resource allocation in multi-{UAV} enabled {IoT} edge network,'' \emph{IEEE Trans. Netw. Serv. Manage.}, vol.~18, no.~4, pp. 4531--4547, 2021.

\bibitem{zheng2022}
G.~Zheng, C.~Xu, M.~Wen, and X.~Zhao, ``Service caching based aerial cooperative computing and resource allocation in multi-uav enabled {MEC} systems,'' \emph{{IEEE} Trans. Veh. Technol.}, vol.~71, no.~10, pp. 10\,934--10\,947, 2022.

\bibitem{Chen2017}
M.~Chen, M.~Mozaffari, W.~Saad, C.~Yin, M.~Debbah, and C.~S. Hong, ``Caching in the sky: {P}roactive deployment of cache-enabled unmanned aerial vehicles for optimized quality-of-experience,'' \emph{{IEEE} J. Sel. Areas Commun.}, vol.~35, no.~5, pp. 1046--1061, 2017.

\bibitem{Wang2016}
Y.~Wang, M.~Sheng, X.~Wang, L.~Wang, and J.~Li, ``Mobile-edge computing: {P}artial computation offloading using dynamic voltage scaling,'' \emph{{IEEE} Trans. Commun.}, vol.~64, no.~10, pp. 4268--4282, 2016.

\bibitem{Gu2021}
Y.~Gu, Y.~Yao, C.~Li, B.~Xia, D.~Xu, and C.~Zhang, ``Modeling and analysis of stochastic mobile-edge computing wireless networks,'' \emph{{IEEE} Internet Things J.}, vol.~8, no.~18, pp. 14\,051--14\,065, 2021.

\bibitem{zhao2023deep}
N.~Zhao, Y.~Pei, Y.-C. Liang, and D.~Niyato, ``Deep reinforcement learning-based contract incentive mechanism for joint sensing and computation in mobile crowdsourcing networks,'' \emph{{IEEE} Internet Things J.}, 2023.

\bibitem{Wang2023}
Z.~Wang, Y.~Wei, Z.~Feng, F.~R. Yu, and Z.~Han, ``Resource management and reflection optimization for intelligent reflecting surface assisted multi-access edge computing using deep reinforcement learning,'' \emph{{IEEE} Trans. Wirel. Commun.}, vol.~22, no.~2, pp. 1175--1186, 2023.

\bibitem{Deng2022}
Y.~Deng, Z.~Chen, X.~Chen, and Y.~Fang, ``Throughput maximization for multiedge multiuser edge computing systems,'' \emph{{IEEE} Internet Things J.}, vol.~9, no.~1, pp. 68--79, 2022.

\bibitem{bando2012}
K.~Bando, ``Many-to-one matching markets with externalities among firms,'' \emph{J. Math. Econ.}, vol.~48, no.~1, pp. 14--20, 2012.

\bibitem{Zhao2017}
J.~Zhao, Y.~Liu, K.~K. Chai, Y.~Chen, and M.~Elkashlan, ``Many-to-many matching with externalities for device-to-device communications,'' \emph{{IEEE} Wirel. Commun. Lett.}, vol.~6, no.~1, pp. 138--141, 2017.

\bibitem{BodineBaron2011}
E.~Bodine{-}Baron, C.~Lee, A.~Chong, B.~Hassibi, and A.~Wierman, ``Peer effects and stability in matching markets,'' in \emph{Proc. {SAGT}}, vol. 6982, 2011, pp. 117--129.

\bibitem{Huo2024}
X.~Huo, H.~Zhang, Z.~Wang, H.~Yan, and C.~Liu, ``An efficient matching game approach to association formation in uav-enabled hierarchical distributed learning,'' \emph{IEEE Trans. Cybern.}, 2024.

\bibitem{Haarnoja2018}
T.~Haarnoja, A.~Zhou, K.~Hartikainen, G.~Tucker, S.~Ha, J.~Tan, V.~Kumar, H.~Zhu, A.~Gupta, P.~Abbeel, and S.~Levine, ``Soft actor-critic algorithms and applications,'' \emph{CoRR}, vol. abs/1812.05905, 2018.

\bibitem{Mu2021}
X.~Mu, Y.~Liu, L.~Guo, J.~Lin, and Z.~Ding, ``Energy-constrained {UAV} data collection systems: {NOMA} and {OMA},'' \emph{{IEEE} Trans. Veh. Technol.}, vol.~70, no.~7, pp. 6898--6912, 2021.

\bibitem{td32018}
S.~Fujimoto, H.~van Hoof, and D.~Meger, ``Addressing function approximation error in actor-critic methods,'' \emph{CoRR}, vol. abs/1802.09477, 2018.

\bibitem{ppo2017}
J.~Schulman, F.~Wolski, P.~Dhariwal, A.~Radford, and O.~Klimov, ``Proximal policy optimization algorithms,'' \emph{CoRR}, vol. abs/1707.06347, 2017.

\bibitem{li2024uav}
W.~Li, S.~Li, H.~Shi, W.~Yan, and Y.~Zhou, ``{UAV}-enabled fair offloading for {MEC} networks: {A} {DRL} approach based on actor-critic parallel architecture,'' \emph{Appl. Intell.}, vol.~54, no.~4, pp. 3529--3546, 2024.

\bibitem{Huang2024}
T.~Huang, Z.~Fang, Q.~Tang, R.~Xie, T.~Chen, and F.~R. Yu, ``Dual-timescales optimization of task scheduling and resource slicing in satellite-terrestrial edge computing networks,'' \emph{{IEEE} Trans. Mob. Comput.}, vol.~23, no.~12, pp. 14\,111--14\,126, 2024.

\end{thebibliography}

\end{document}